\documentclass[conference]{IEEEtran}
\IEEEoverridecommandlockouts
\usepackage{amsmath,amssymb,amsfonts}
\usepackage{algorithm,algpseudocode}
\usepackage{cite}
\usepackage{hyperref}
\usepackage{makecell}
\usepackage{multirow}
\usepackage{graphicx}
\usepackage{textcomp}
\usepackage{url}
\usepackage{xcolor}
\def\BibTeX{{\rm B\kern-.05em{\sc i\kern-.025em b}\kern-.08em
    T\kern-.1667em\lower.7ex\hbox{E}\kern-.125emX}}
\setcellgapes{3pt}
\begin{document}

\title{Building a 3-Player Mahjong AI using\\Deep Reinforcement Learning } 

\author{\IEEEauthorblockN{Xiangyu Zhao}
\IEEEauthorblockA{\textit{Department of Computer Science and Technology} \\
\textit{University of Cambridge}\\
Cambridge, United Kingdom \\
xz398@cam.ac.uk}
\and
\IEEEauthorblockN{Sean B. Holden}
\IEEEauthorblockA{\textit{Department of Computer Science and Technology} \\
\textit{University of Cambridge}\\
Cambridge, United Kingdom \\
sbh11@cl.cam.ac.uk}}


\maketitle

\begin{abstract}
Mahjong is a popular multi-player imperfect-information game developed in China in the late 19th-century, with some very challenging features for AI research. Sanma, being a 3-player variant of the Japanese Riichi Mahjong, possesses unique characteristics including fewer tiles and, consequently, a more aggressive playing style. It is thus challenging and of great research interest in its own right, but has not yet been explored. In this paper, we present Meowjong, an AI for Sanma using deep reinforcement learning. We define an informative and compact 2-dimensional data structure for encoding the observable information in a Sanma game. We pre-train 5 convolutional neural networks (CNNs) for Sanma's 5 actions---discard, Pon, Kan, Kita and Riichi, and enhance the major action's model, namely the discard model, via self-play reinforcement learning using the Monte Carlo policy gradient method. Meowjong's models achieve test accuracies comparable with AIs for 4-player Mahjong through supervised learning, and gain a significant further enhancement from reinforcement learning. Being the first ever AI in Sanma, we claim that Meowjong stands as a state-of-the-art in this game.
\end{abstract}

\begin{IEEEkeywords}
Mahjong, deep learning, reinforcement learning, convolutional neural networks, policy gradient methods
\end{IEEEkeywords}

\section{Introduction}

Mahjong is a popular tile-based multi-round multi-player imperfect-information game that was developed in China in the late 19th century, and nowadays has hundreds of millions of players worldwide. It is a game of skill, strategy and calculation, and involves a degree of chance. Mahjong is very challenging for AI because: 

\begin{itemize}
	\item Mahjong is played by more than 2 players and has significant hidden information;
	\item Mahjong has complex playing and scoring rules;
	\item Mahjong has a huge number of possible winning hands in various patterns, allowing flexible in-game strategy adaptation.
\end{itemize}

While there are many variants of Mahjong, we focus on Sanma, a 3-player variant of the Japanese Riichi Mahjong. Sanma differs fundamentally from 4-player Japanese Riichi Mahjong because most of the tiles in a suit (Manzu) are removed, and one of the actions (Chii) is replaced with another action (Kita). Consequently, both the number of possible states and the amount of hidden information are reduced, and hands tend to develop much faster. As a result, players tend to play more aggressively, and valuable winning hands become more frequent. Since it is commonly understood by expert human players that the 4-player Japanese Riichi Mahjong should be played defensively, an AI trained to learn optimal strategy for the 4-player Japanese Riichi Mahjong is not guaranteed to acquire the optimal strategy for Sanma. Therefore, even though Sanma is seemingly simpler than 4-player Mahjong, it still preserves all the challenging characteristics of Mahjong, and the difference in strategy also makes it a worthwhile target for research.

In this paper, we describe Meowjong, a Sanma AI based on deep reinforcement learning (DRL). First, we propose a 2-dimensional data structure for encoding the observable information in a round of a Sanma game. We then pre-train 5 deep convolutional neural networks (CNNs), each corresponding to one action in Sanma (discard, Pon, Kan, Kita, and Riichi), and improve the discard model using the Monte Carlo policy gradient method for self-play reinforcement learning (RL).

\section{Related Works}

As far as we are aware, there has not been any published attempt to develop an AI player for Sanma. Various machine learning approaches have been tried on 4-player Mahjong, including Bakuuchi by Mizukami et al. \cite{Mizukami2015Bakuuchi} in 2015, which was based on Monte Carlo simulation and opponent modelling. Bakuuchi has achieved a rating of 1718 on Tenhou \cite{Tenhou} (a popular global online Riichi Mahjong platform with more than 350,000 active users); this was significantly higher than that of an average human player. In 2019, Kurita et al. \cite{Kurita2021Mahjong} built a Mahjong AI based on multiple Markov decision processes as abstractions of Mahjong, and also reached the top level of Mahjong AI, as demonstrated by playing directly against Bakuuchi. In 2018, Gao et al. \cite{Gao2018Mahjong} built a Mahjong AI using CNNs and achieved a test accuracy of 68.8\% for the discard action, which is 6.7\% higher than the previous best test accuracy of 62.1\% reported by Bakuuchi. They also evaluated the AI's strength on Tenhou, reaching a rating of 1822 after 76 matches. In 2019, Li et al. \cite{Li2020Suphx} built a Mahjong AI called Suphx, based on DRL. Suphx parameterized its policy using CNNs, and improved its discard model through distributed RL. It also introduced global reward prediction based on gated recurrent units (GRUs) to predict the final reward of an entire game based on the information of the current and previous rounds. Furthermore, it introduced oracle guiding to speed up the agent's improvement in RL. During online playing, Suphx employed run-time parametric Monte Carlo policy adaptation to exploit the new observations on the current round in order to perform even better. Suphx eventually reached a stable rank of 8.74 dan on Tenhou, which is about 2 dan higher than Bakuuchi, and is higher than 99.99\% of all the officially ranked human players on Tenhou, though at a cost of needing extremely heavy computational resources for training. While the performance of these systems is impressive, we emphasize that, as we are unaware of comparable work addressing Sanma, no such comparisons with human players are available in this context.

\section{Basic Rules and Terminology of Sanma}

Sanma has 108 \textit{tiles} of 27 different kinds, with four identical tiles in each kind. TABLE~\ref{table:mahjong-tiles} shows the appearances of the 27 kinds of tiles. These 27 kinds are:

\begin{itemize}
	\item \textit{Manzu}: 1m (Man) and 9m, with traditional Chinese character patterns;
	\item \textit{Pinzu}: 1--9p (Pin), with dot patterns;
	\item \textit{Souzu}: 1--9s (Sou), with bamboo patterns (1s is often decorated with a bird);
	\item \textit{Honours}: this includes four \textit{winds}: East, South, West, North, and three \textit{dragons}: Haku (white), Hatsu (green), Chun (red). 
\end{itemize}

At the start of each round, the tiles are shuffled and arranged into \textit{walls} of face-down two-tier-high tiles. Each player receives 13 tiles from the wall as their starting \textit{hand}. The last 14 tiles form the \textit{dead wall}, which will not be used throughout the current round except in some specific cases, and the rest of the walls with available tiles form the \textit{live walls}. The seats of the players are denoted East, South and West, also referred to as the players' \textit{seat winds}. The East player in each round becomes the dealer. In each round, the players take turns to draw and discard tiles until one of them completes a winning hand. The round ends with a draw if all tiles are drawn and discarded. Other than the normal discard action, a player can perform the following special actions in legal situations, which may interrupt the regular playing order:

\begin{itemize}
	\item \textit{Pon}: a player can claim an immediately discarded tile from any other player to form a \textit{triplet} of the same kind (also called a \textit{Koutsu}). This action may cause a player's turn to be skipped. The player needs to discard a tile after calling Pon.
	\item \textit{Kan}: a player can call Kan to form a \textit{quad} of the same kind (also called a \textit{Kantsu}). The player draws another tile and continues their turn after a Kan, and a \textit{Dora indicator} is also revealed from the dead wall. A Dora indicator indicates corresponding \textit{Doras}; these are bonus tiles that add value to a winning hand.
	\item \textit{Kita}: a player can put a North tile to the side of their hand, and it would be counted as \textit{Dora}, called the \textit{Nukidora}. The player is awarded an extra draw and can continue their turn. 
	\item \textit{Riichi}: a player can pay a 1,000 point deposit and declare a ready hand, which means they only need one more tile to win. After declaring Riichi, the player's hand cannot be changed, and the player must discard any tile they draw until the round ends. The Riichi player bares certain risks, but has an enhanced chance to win and can gain a larger score if they do.
\end{itemize}

Knowing when to appropriately make those actions is one of the central strategies in Riichi Mahjong.

There are two types of winning: winning from another player's discard (\textit{Ron}), or winning from self-draw (\textit{Tsumo}). In the case of Ron, the player who feeds the winning tile pays the winning player by the amount of the winning player's hand score. In the case of Tsumo, the other two players split the winning player's hand score and pay the winning player together. The dealer wins 50\% more, but when a non-dealer player wins by Tsumo, the dealer also pays twice compared to the other paying player.

\begin{table}[t]
	\makegapedcells
	\centering
	\caption{Mahjong tiles$^{\mathrm{a}}$}
	\renewcommand{\arraystretch}{1.1}
	\resizebox{\columnwidth}{!}{\begin{tabular} {|c|c|c|c|c|c|c|c|c|c|}
		\hline
		\multirow{2}{*}{ } & \multicolumn{9}{c|}{\textbf{Numbers}} \\
		\cline{2-10}
		& \textbf{\textit{1}} & \textbf{\textit{2}} & \textbf{\textit{3}} & \textbf{\textit{4}} & \textbf{\textit{5}} & \textbf{\textit{6}} & \textbf{\textit{7}} & \textbf{\textit{8}} & \textbf{\textit{9}} \\
		\hline
		Manzu & \includegraphics[width=0.06\columnwidth]{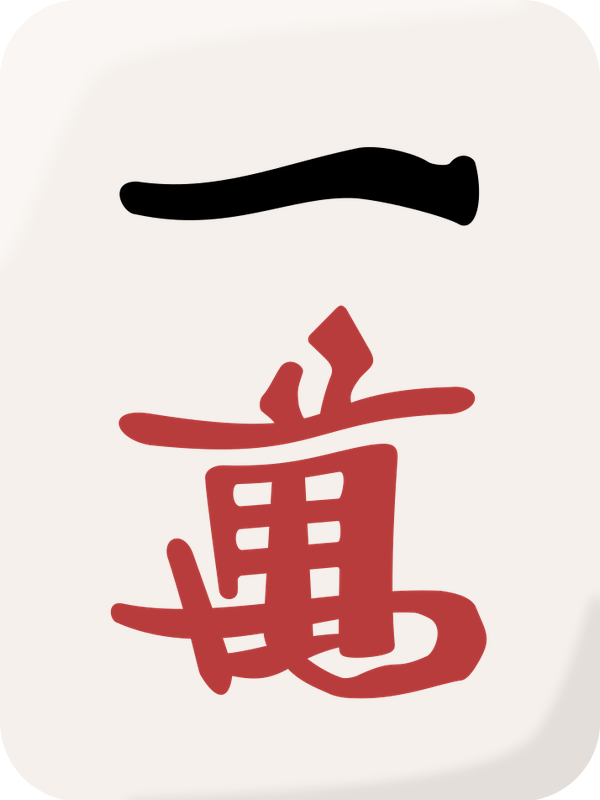} & {} & {} & {} & {} & {} & {} & {} & \includegraphics[width=0.06\columnwidth]{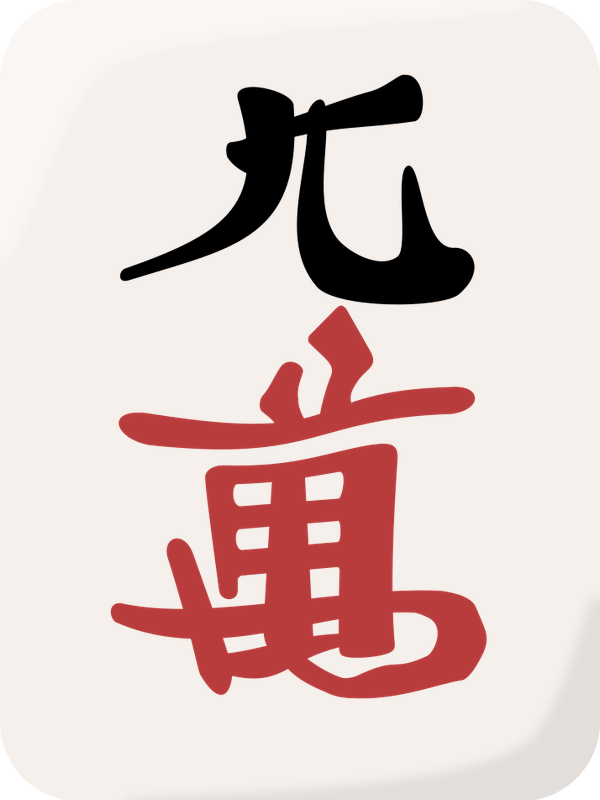} \\
		\hline
		Pinzu & \includegraphics[width=0.06\columnwidth]{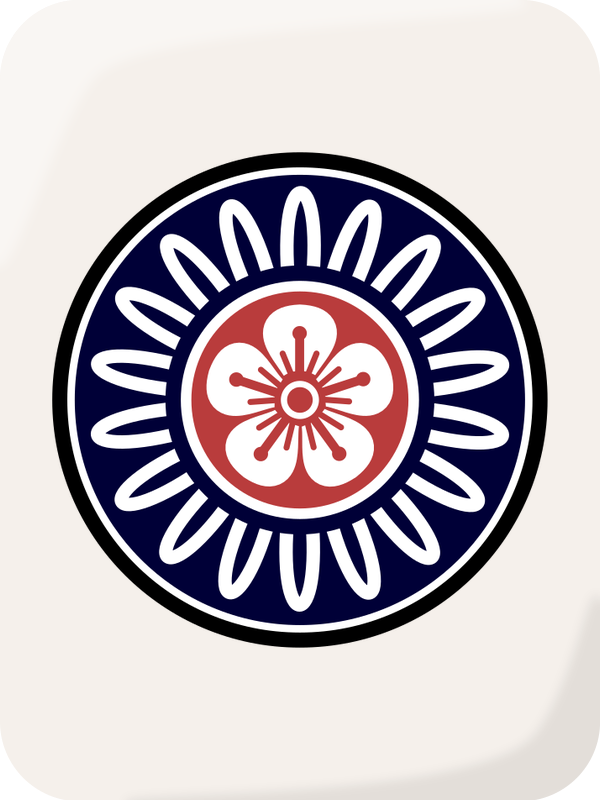} & \includegraphics[width=0.06\columnwidth]{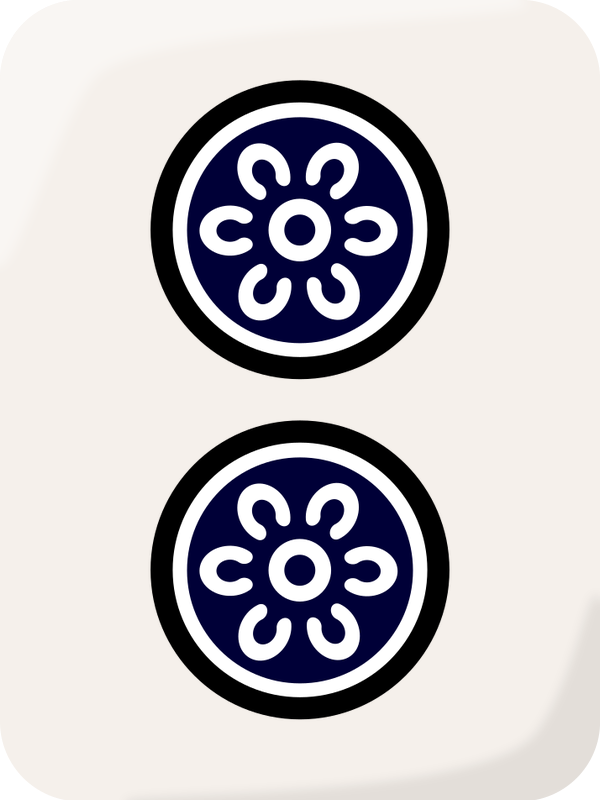} & \includegraphics[width=0.06\columnwidth]{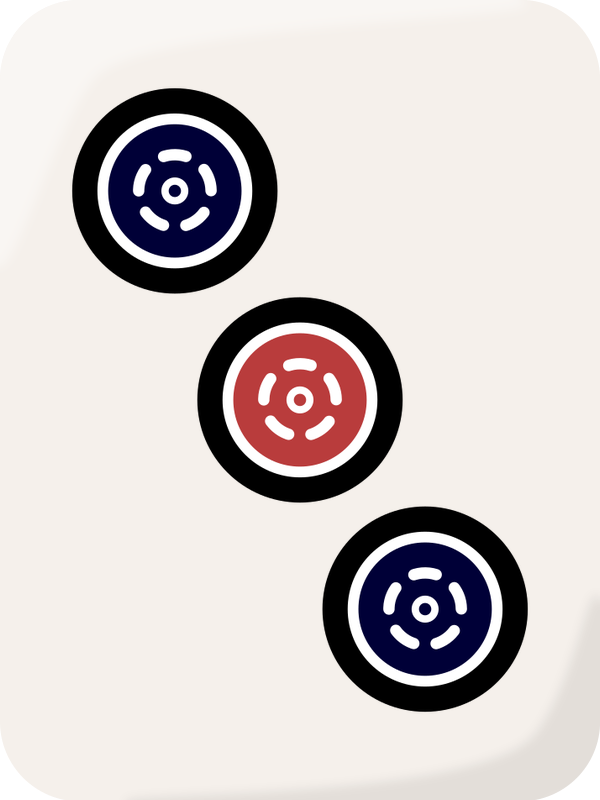} & \includegraphics[width=0.06\columnwidth]{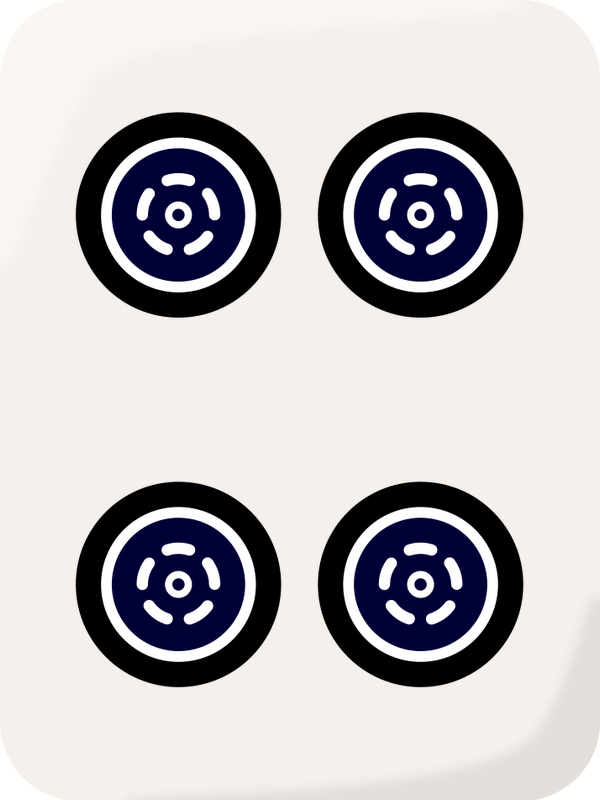} & \includegraphics[width=0.06\columnwidth]{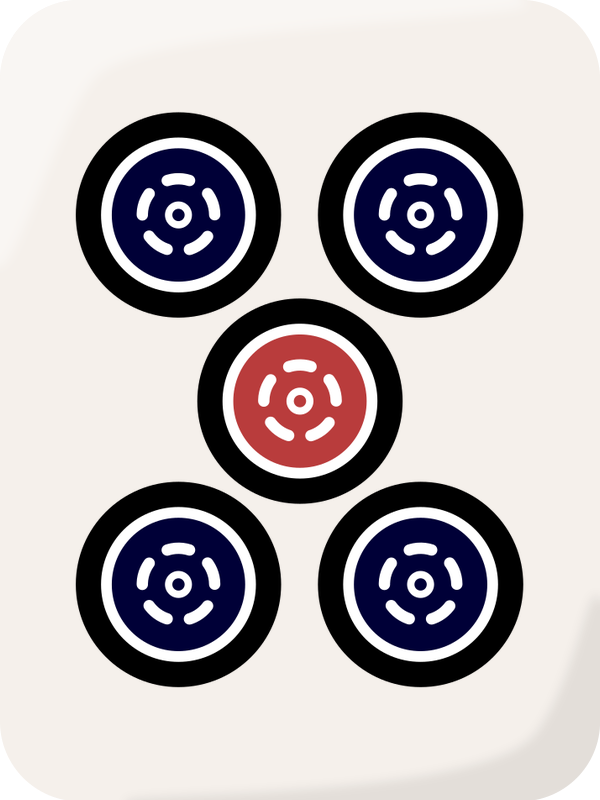} & \includegraphics[width=0.06\columnwidth]{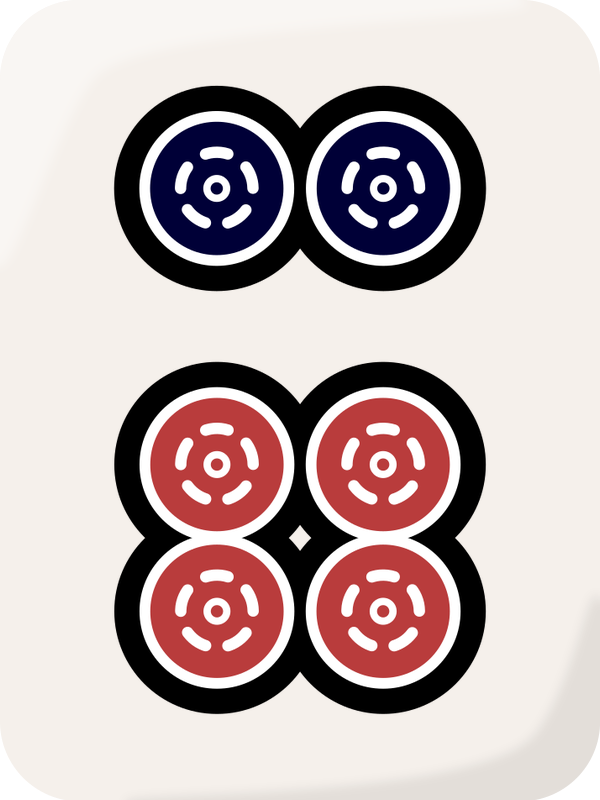} & \includegraphics[width=0.06\columnwidth]{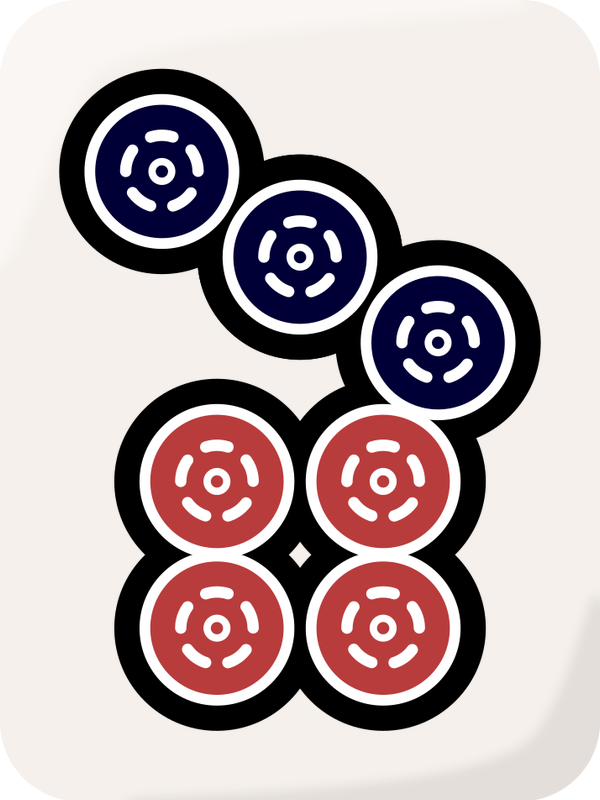} & \includegraphics[width=0.06\columnwidth]{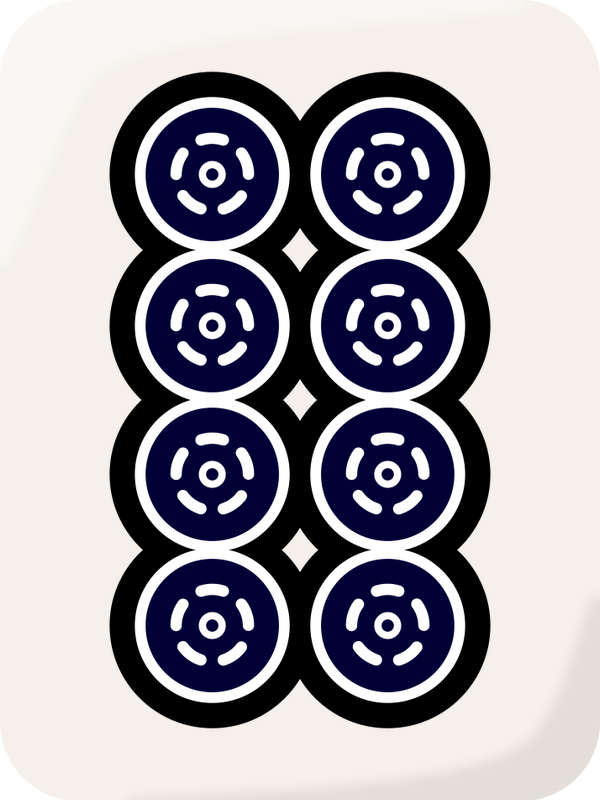} & \includegraphics[width=0.06\columnwidth]{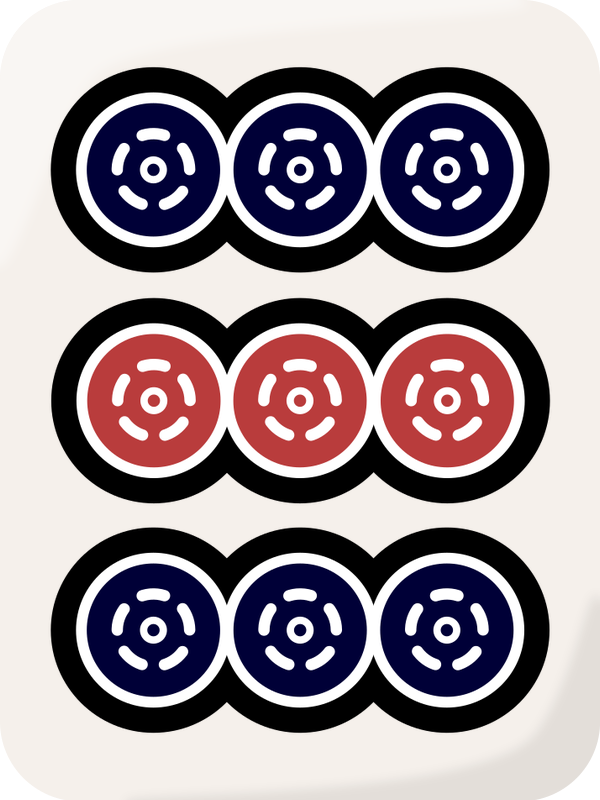} \\
		\hline
		Souzu & \includegraphics[width=0.06\columnwidth]{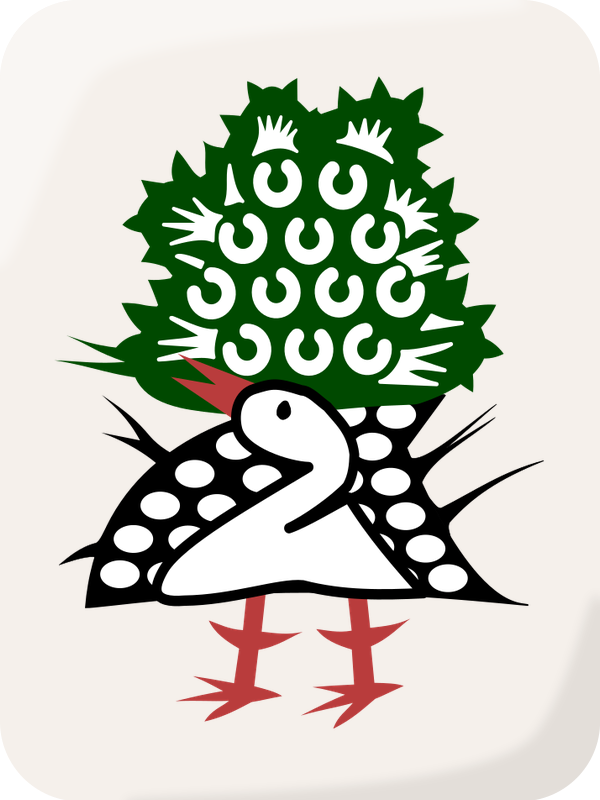} & \includegraphics[width=0.06\columnwidth]{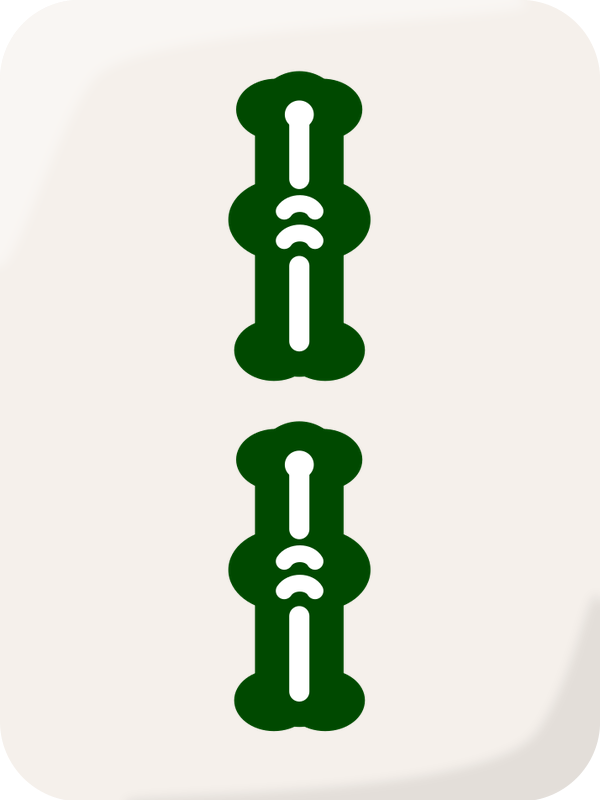} & \includegraphics[width=0.06\columnwidth]{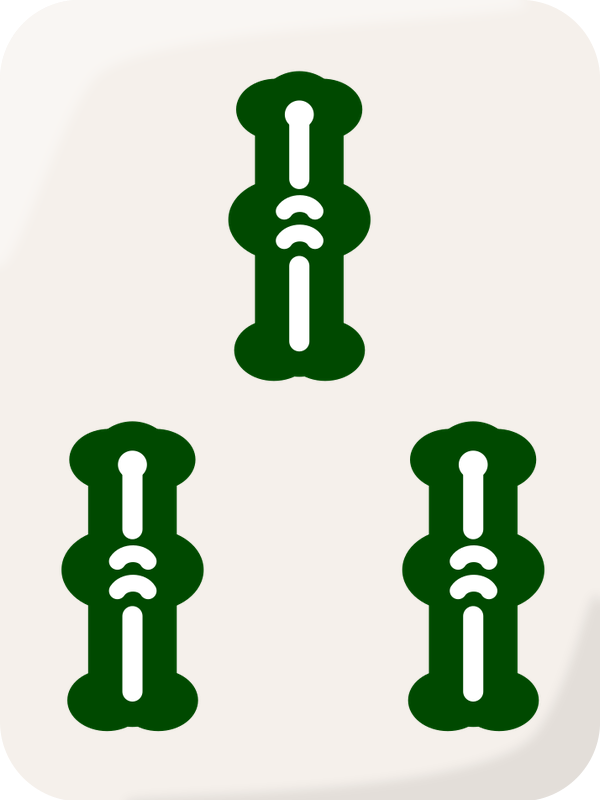} & \includegraphics[width=0.06\columnwidth]{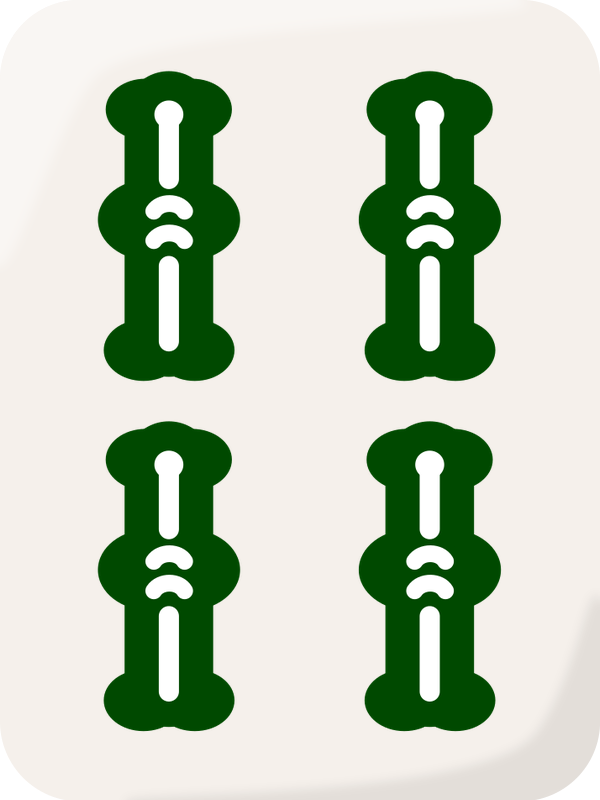} & \includegraphics[width=0.06\columnwidth]{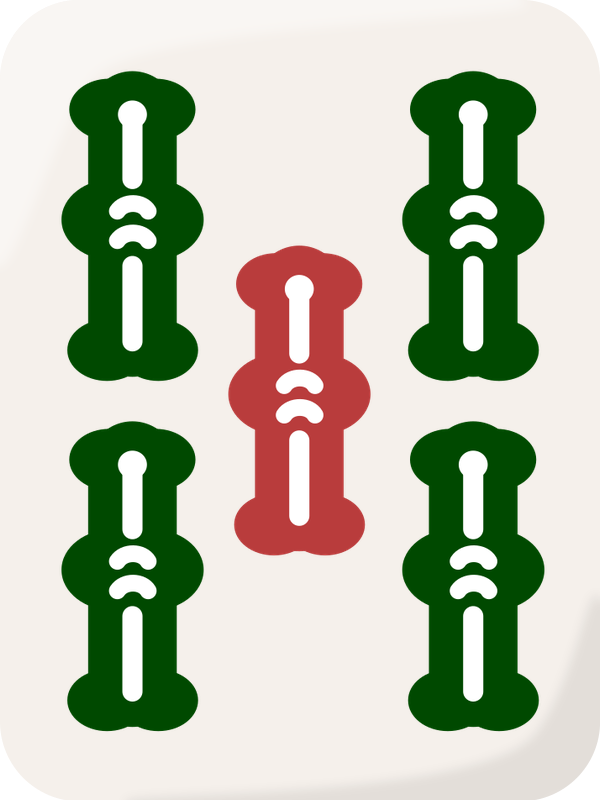} & \includegraphics[width=0.06\columnwidth]{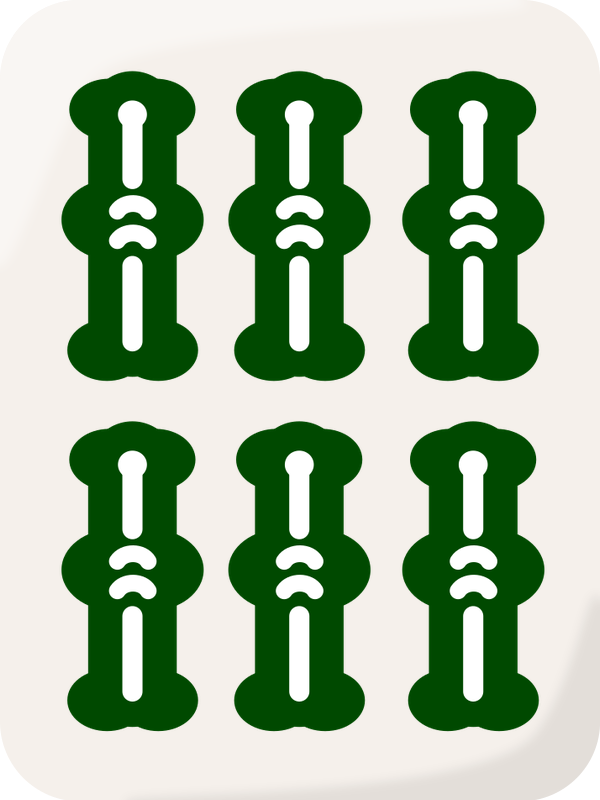} & \includegraphics[width=0.06\columnwidth]{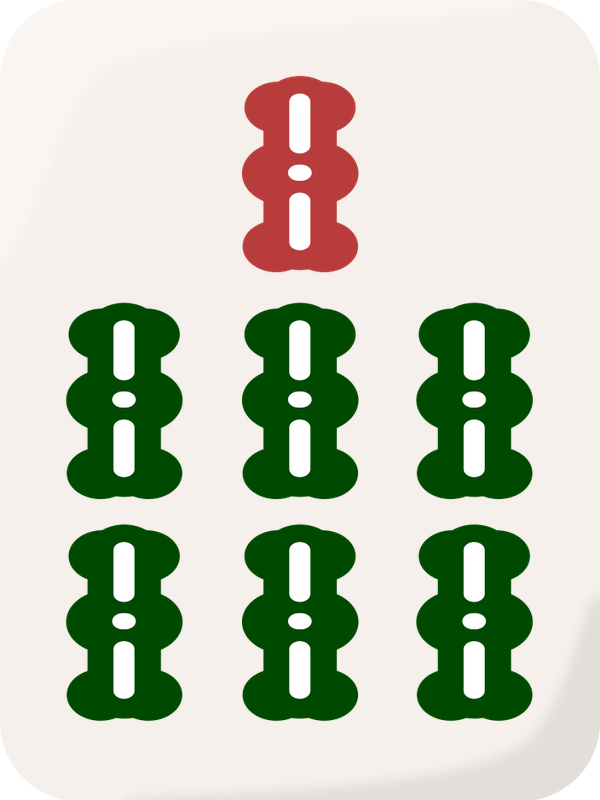} & \includegraphics[width=0.06\columnwidth]{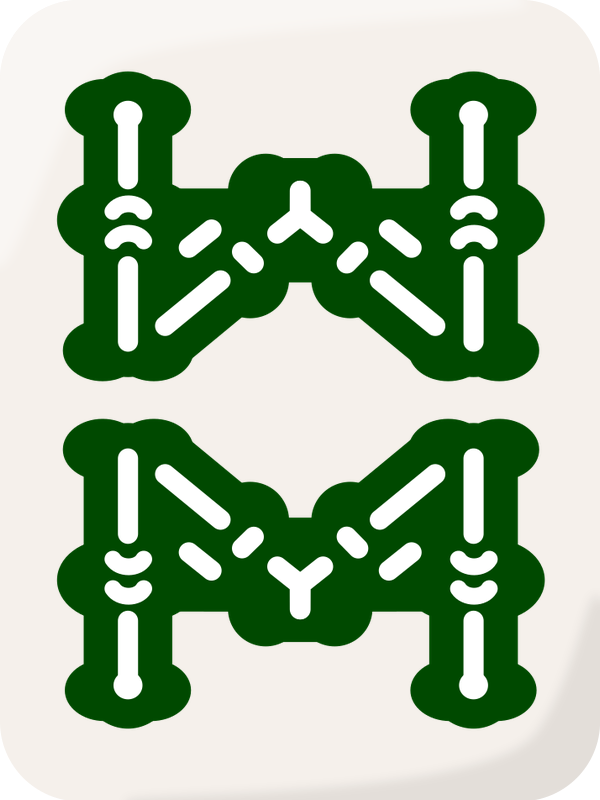} & \includegraphics[width=0.06\columnwidth]{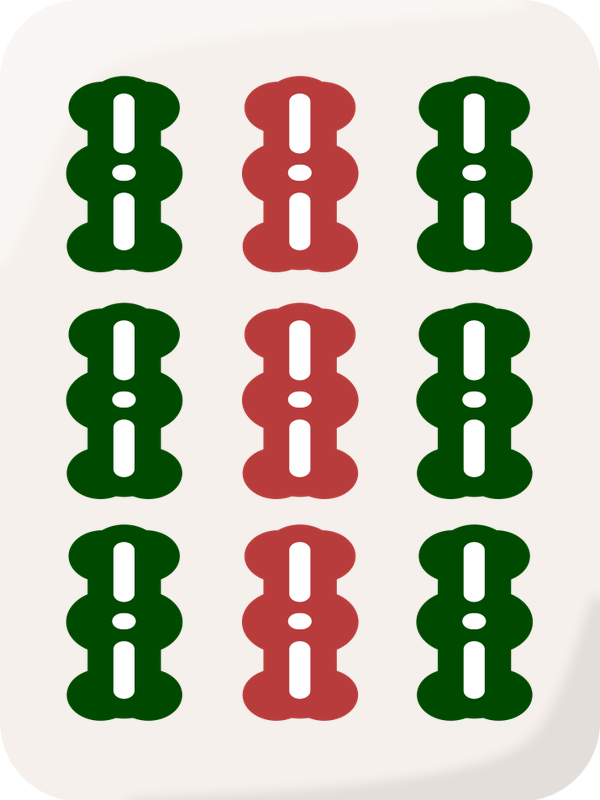} \\
		\hline
		\multirow{3}{*}{\vspace*{-5ex}Honours} & \multicolumn{4}{c|}{Winds} & \multicolumn{3}{c|}{Dragons} & \multicolumn{2}{|c}{\multirow{3}{*}{ }} \\
		\cline{2-8}
		& East & South & West & North & Haku & Hatsu & Chun & \multicolumn{2}{|c}{ } \\
		\cline{2-8}
		& \includegraphics[width=0.06\columnwidth]{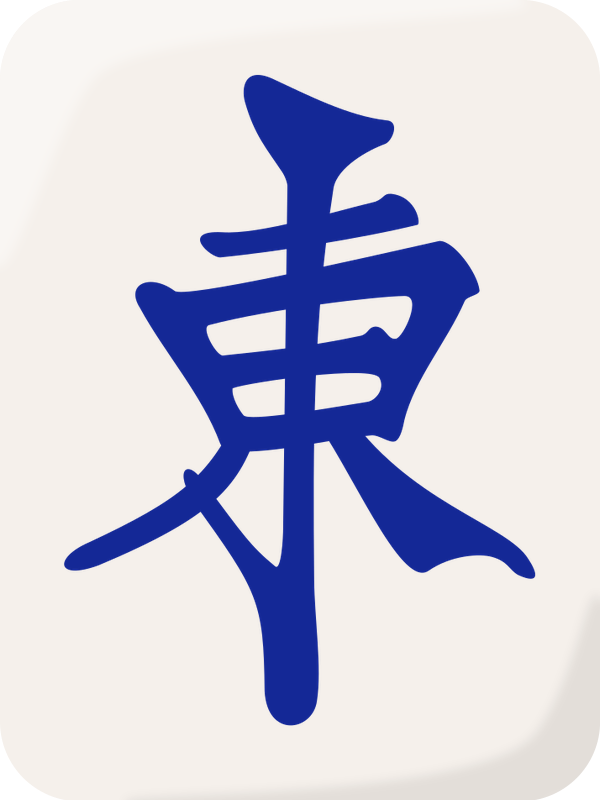} & \includegraphics[width=0.06\columnwidth]{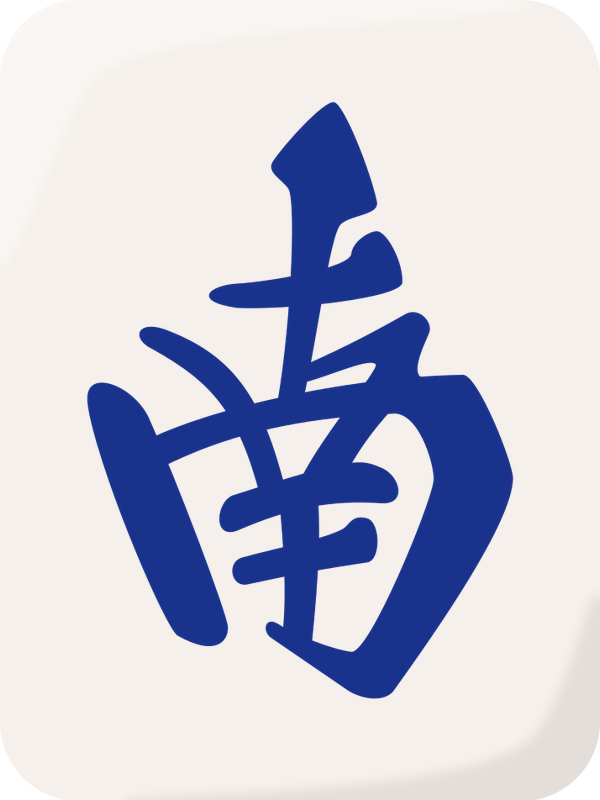} & \includegraphics[width=0.06\columnwidth]{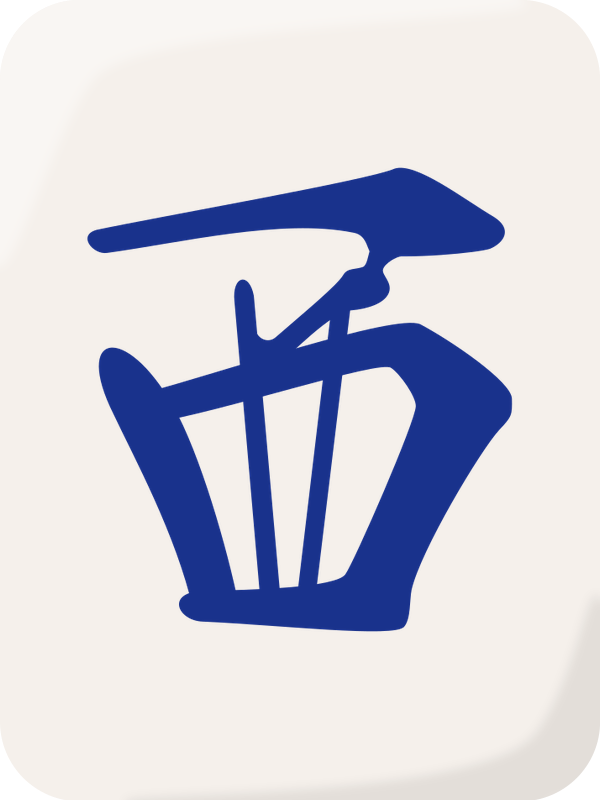} & \includegraphics[width=0.06\columnwidth]{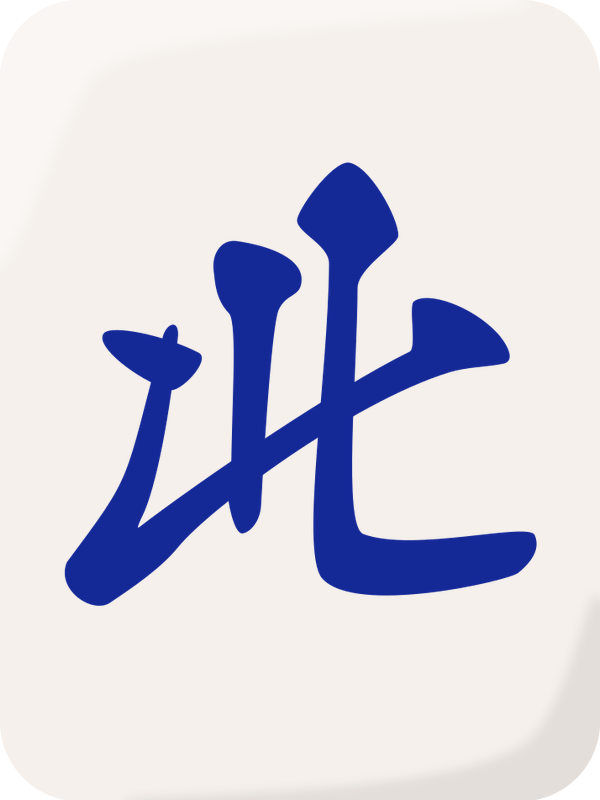} & \includegraphics[width=0.06\columnwidth]{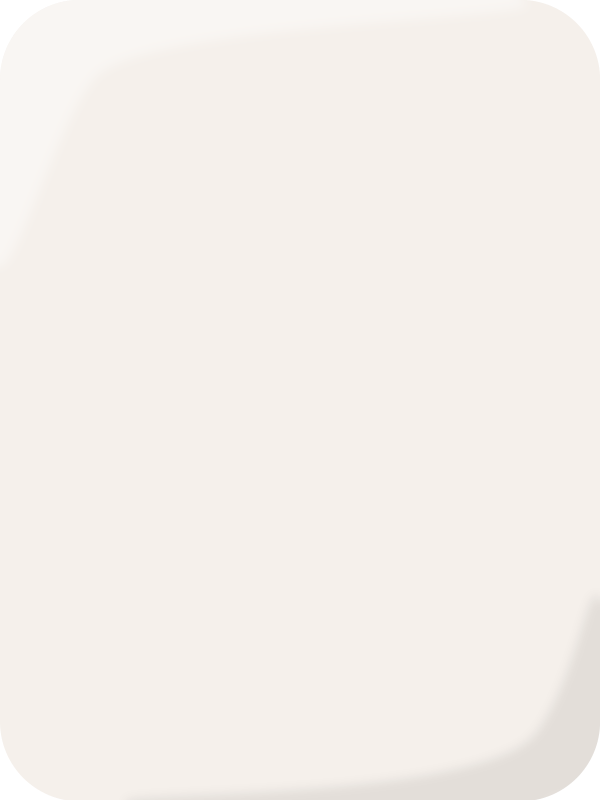} & \includegraphics[width=0.06\columnwidth]{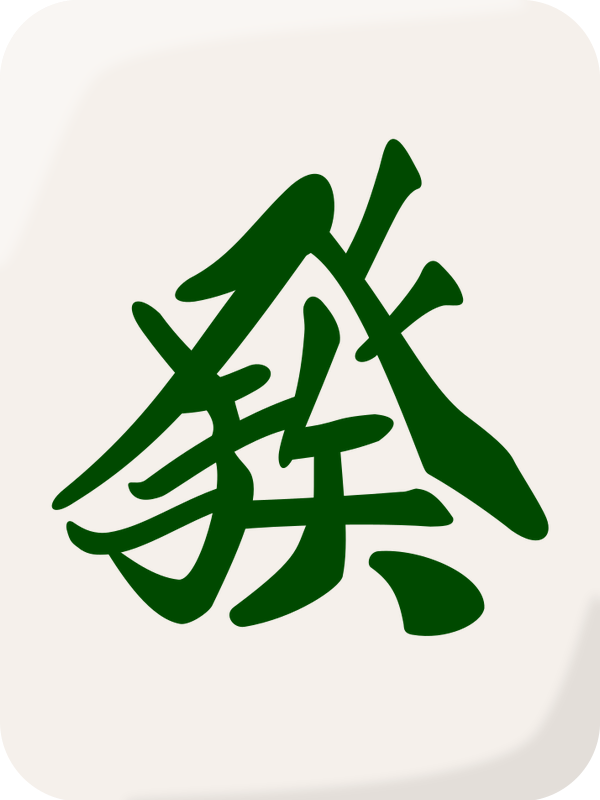} & \includegraphics[width=0.06\columnwidth]{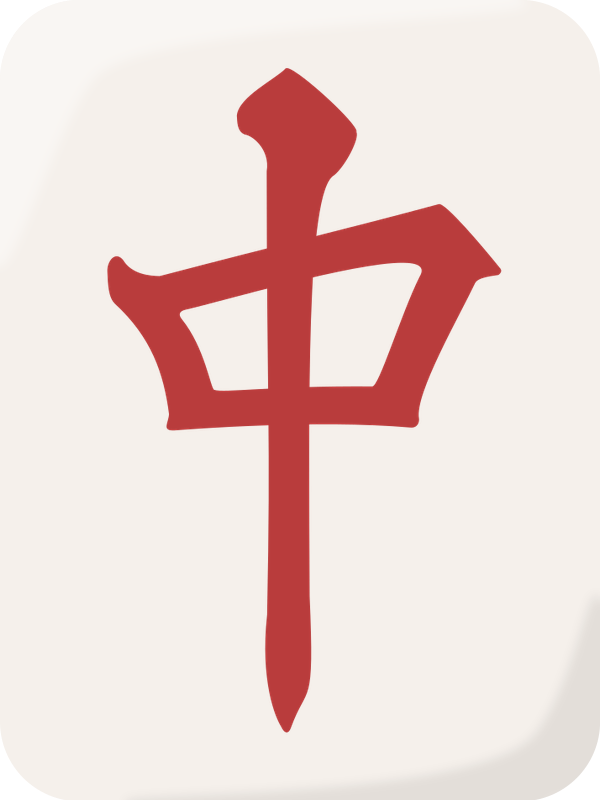} & \multicolumn{2}{|c}{ } \\
		\cline{1-8}
		\multicolumn{10}{l}{$^{\mathrm{a}}$ Graphical resources adapted from (\url{https://github.com/FluffyStuff/riichi-mahjong-tiles})} \\[-0.5\baselineskip]
		\multicolumn{10}{l}{\phantom{$^{\mathrm{a}}$ }under the Creative Commons Attribution (CC BY) licence} \\[-0.5\baselineskip]
		\multicolumn{10}{l}{\phantom{$^{\mathrm{a}}$ }(\url{https://creativecommons.org/licenses/by/4.0/}).} \\
	\end{tabular}}
	\label{table:mahjong-tiles}
\end{table}

\section{Features and Data Structure Design}

Unlike other board games such as Chess and Go, the layout of a Mahjong board is not standardized, and the observable information must be carefully encoded in order to be digested by the CNNs. Since there are 34 different tiles in Mahjong, we use a $34\times366$ array to represent a state. Although 2m--8m are not included in Sanma, which leaves only 27 of the 34 tiles to be used, we still include all the tiles and left the excluded tiles blank, to add transferability of Meowjong to 4-player Mahjong. The mapping between the tiles and their corresponding row indices in the array encoding is shown in TABLE~\ref{table:tiles-indices-mapping}. 

\begin{figure*}[t]
	\centering
	\includegraphics[width=0.67\textwidth]{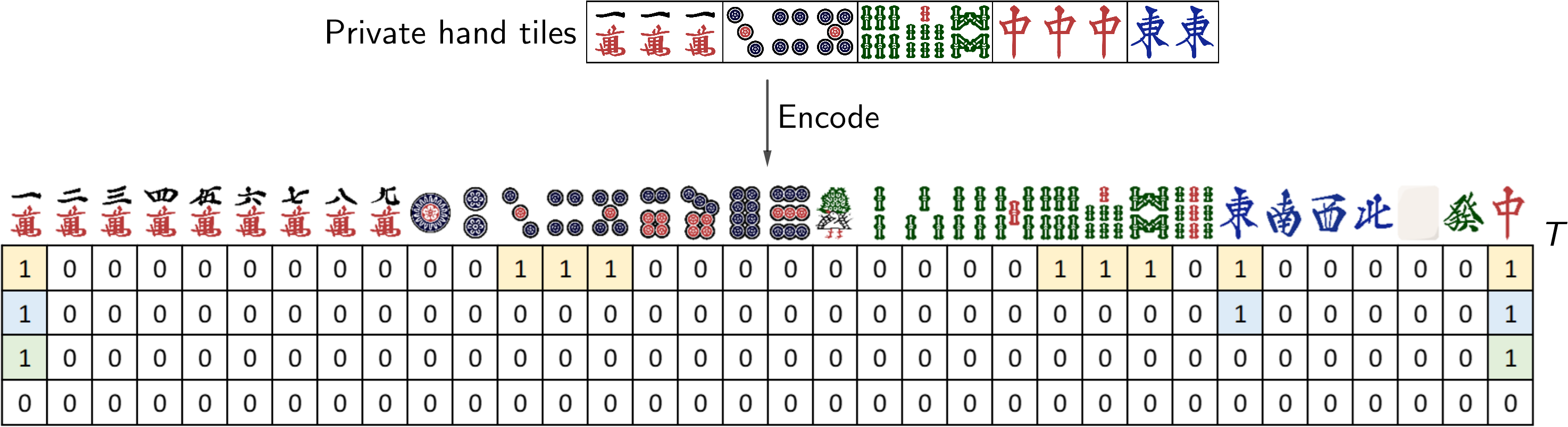}
	\caption{Encoding of an example private hand using 4 channels (transposed to save space)}
	\label{figure:tiles-encoding}
\end{figure*}

To simulate the real in-game environment and maximize Meowjong's performance, all observable information should be taken into account. Therefore, we use 366 columns in total to represent 22 features, as listed in TABLE~\ref{table:features}. We do not include the number of remaining tiles, because it can be calculated from the number of discarded tiles. For triplets, quads and Riichi status, we include not only the triplet/quad tiles and the Riichi status, but also the turn numbers of the Pon/Kan/Riichi calls. Besides, although the spaces for the fourth player are not needed for Sanma, they are still created for the sake of transferability of Meowjong to the 4-player Mahjong.

\begin{table}[t]
	\centering
	\caption{Tiles and their corresponding row index representations}
	\resizebox{0.9\columnwidth}{!}{\begin{tabular}{|l|c|}
		\hline
		\textbf{Tiles} & \textbf{Corresponding indices} \\
		\hline
		Manzu, i.e., 1m--9m & 0--8 \\
		Pinzu, i.e., 1p--9p & 9--17 \\
		Souzu, i.e., 1s--9s & 18--26 \\
		Winds, i.e., East, South, West, and North & 27--30 \\
		Honours, i.e., Haku, Hatsu, and Chun & 31--33 \\
		\hline
	\end{tabular}}
	\label{table:tiles-indices-mapping}
\end{table}

\begin{table}[t]
	\centering
	\caption{Input features for the models}
	\resizebox{0.9\columnwidth}{!}{\begin{tabular}{|l|c|}
		\hline
		\textbf{Feature} & \textbf{Number of channels} \\
		\hline
		Target tile & 1 \\
		Self private tiles & 4 \\
		Red Doras\footnotemark in hand & 1 \\
		Self open triplets/quads & $(4+5)\times4=36$ \\
		Self Kitas & 4 \\
		Self discards & 30 \\
		Dora indicators & 5 \\
		Other players' Riichi status & $(1+5)\times3=18$ \\
		Scores & $11\times4=44$ \\
		Round (Kyoku) number & 4 \\
		Repeats count (i.e., Honba number) & 4 \\
		Deposit count & 4 \\
		Self wind & 1 \\
		Other players' open tiles\footnotemark and discards & $(36+4+30)\times3=210$ \\
		\hline
		Total & 366 \\
		\hline
	\end{tabular}}
	\label{table:features}
\end{table}

\footnotetext[1]{In Mahjong, one of each of the fives in Manzu, Pinzu and Souzu are marked red and count as Doras, called the \textit{red doras} (also known as the \textit{Akadoras}). They are still the same tiles as the ordinary fives, except that they are worth extra points in the score calculation.}
\footnotetext[2]{Triplets, quads, and Kitas.}

The included features can be divided into two categories:

\begin{itemize}
	\item \textit{Tile features} that involve sets or sequences of tiles, such as private tiles, triplets/quads, and so on. Tile features can be encoded by setting the corresponding row indices to 1 and leaving the rest to 0, as, for example, shown in Fig.~\ref{figure:tiles-encoding}.
	\item \textit{Numerical features}, for example, player scores. Numerical features can be binary-encoded into multiple columns, each being either all zeros or all ones.
\end{itemize}

\section{Neural Network Structure Design}

\begin{figure*}[t]
	\centering
	\includegraphics[width=0.6\textwidth]{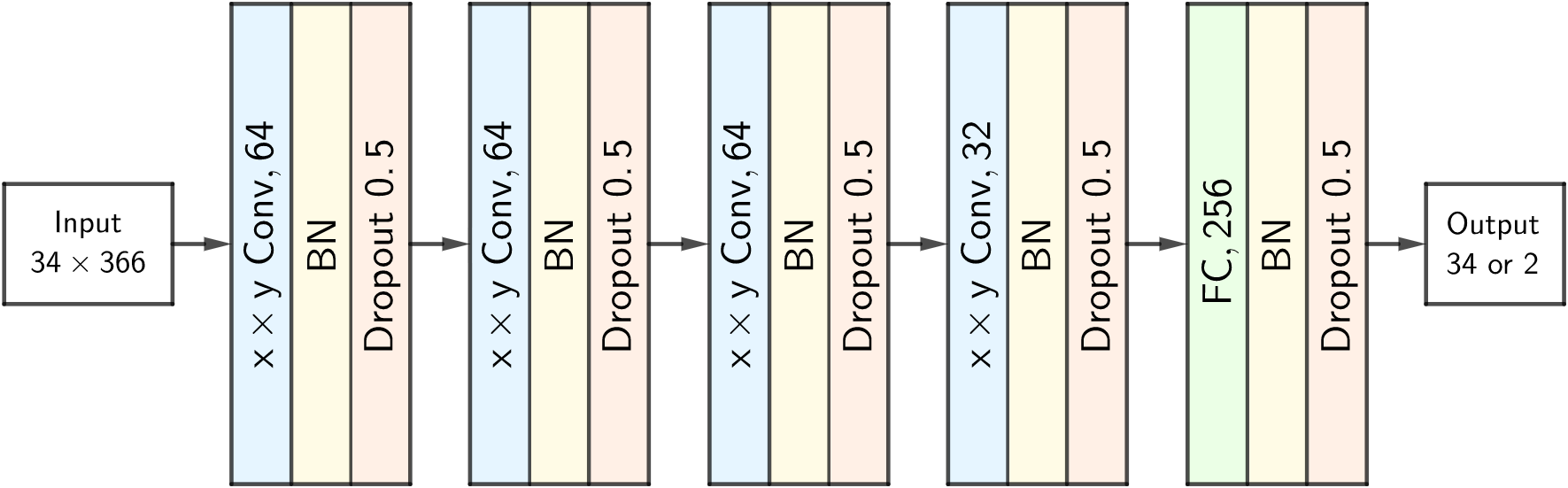}
	\caption{CNN structure of the action models ($x\times y$ denotes the filter size)}
	\label{fig:cnn-structure}
\end{figure*}

We parameterize Meowjong's policy on an action basis: we use a separate model, called an \textit{action model}, to parameterize Meowjong's policy for each action (discard, Pon, Kan, Kita, and Riichi). For the action models of Meowjong, we adopt a CNN structure with 4 convolutional layers followed by a fully-connected layer. Each of the first 3 convolutional layers has 64 filters, and the last convolutional layer has 32 filters. We have experimented with various choices for the model structures, including the number of hidden layers and the number of filters in each layer, and found that having more convolutional layers or more filters per layer could result in an oversized model under our memory constraint. Having fewer convolutional layers would result in a huge number of parameters in the flatten operation before the fully-connected layer, also leading to an oversized model. All filters in the 4 layers share the same size, which is a hyperparameter to be tuned individually for each action model. The fully-connected layer contains 256 hidden nodes. A batch normalization (BN) layer and a dropout layer with dropout rate 0.5 are added after each convolutional and fully-connected layer, in order to prevent over-fitting. All action models share a similar structure, differing in the filter sizes and the output dimensions (34 for the discard network, and 2 for the rest). ReLU is used for the activation function of all the convolutional and fully-connected layers, and softmax is used for the activation function of the output layers. The CNN structure for the action models is shown in Fig.~\ref{fig:cnn-structure}. The input and output dimensions of the models are shown in TABLE~\ref{table:input-output-dimensions}.

Since the decision-making problems in Sanma can be converted to multi-class classification problems, we define the loss for our CNNs to be the categorical cross-entropy
\begin{equation}
	L(\mathbf{w})=-\sum_{i=1}^m\sum_{k=1}^{K}y^{(i)}_k\cdot\log\hat{y}^{(i)}_k
\end{equation}
where $m$ is the number of examples, and $K$ is the number of classes. Although pooling is recognized as an efficient down-sampling tool for CNNs in computer vision tasks due to the shift-invariance property of image recognition, in Meowjong's case, the data structure is not an image, but a compact encoding of discrete feature data, and we would expect the use of pooling here to lose too much information, leading to a lower accuracy. Therefore, pooling is not used in Meowjong's CNN structure. No padding is used in any convolutional layer.

The discard problem can be interpreted as a 34-class classification problem, since there are 34 kinds of tile in total, or a 14-class classification problem, since each player can have at most 14 tiles in their private hand in any given state. As recommended by both Li et al. \cite{Li2020Suphx} and Gao et al. \cite{Gao2018Mahjong}, we adopt the 34-class classification interpretation, and hence use a 34-dimensional output. The only potential risk of the 34-class method is of illegal discards, where the discard model outputs a tile that does not exist in the player's private hand. However, in practice all discard choices made by our discard model are legal; this suggests that our CNN has strong learning ability. For all the rest of the actions, a player can either declare or skip that action in appropriate situations, and all those actions can be interpreted as binary classification problems. Hence, we set up a 2-dimensional output for each of their action models.

Hyperparameter tuning for the pre-training of the CNNs is focused on the filter size for each action model. As there is no guarantee that the same filter size can work for every action, the filter sizes are tuned individually for each action model. Grid search is used for hyperparameter tuning: for each action, all candidate filter sizes $(x,y)$ ranging from $2\leq x,y\leq5$ are tried, making 16 candidate models in total. The best-performing filter sizes for the action models are shown in TABLE~\ref{table:filter-sizes}. Further details are decribed in Section~\ref{sec:experiments}.

\begin{table}[t]
	\centering
	\caption{Input and output dimensions of the action models}
	\resizebox{0.75\columnwidth}{!}{\begin{tabular}{|c|c|c|c|c|c|}
			\hline
			& \textbf{Discard} & \textbf{Pon} & \textbf{Kan} & \textbf{Kita} & \textbf{Riichi} \\
			\hline
			Input & \multicolumn{5}{c|}{$34\times366$} \\
			\hline
			Output & 34 & \multicolumn{4}{c|}{2} \\
			\hline
	\end{tabular}}
	\label{table:input-output-dimensions}
\end{table}

\begin{table}[t]
	\centering
	\caption{filter sizes $(x, y)$ of the action models}
	\resizebox{0.9\columnwidth}{!}{\begin{tabular}{|c|c|c|c|c|c|}
		\hline
		& \textbf{Discard} & \textbf{Pon} & \textbf{Kan} & \textbf{Kita} & \textbf{Riichi} \\
		\hline
		Filter size $(x, y)$ & $(4,5)$ & $(5,4)$ & $(2,3)$ & $(3,2)$ & $(3,4)$ \\
		\hline
	\end{tabular}}
	\label{table:filter-sizes}
\end{table}

\section{Reinforcement Learning}

For RL training of Meowjong, we adopt the Monte Carlo policy gradient method: at each time step $t=0,1,2,\cdots,T$ of a round of Sanma, our agent receives a representation of the game state $S_t$, and on that basis selects an action $A_t$. One time step later, as a consequence of its action, the agent receives a numerical reward $R_{t+1}$ and finds itself in a new state $S_{t+1}$. The agent's goal is to maximize the cumulative reward
\begin{equation}
	\label{eqn:cumulative-reward}
	\begin{aligned}
		G_t &= R_{t+1}+\gamma R_{t+2}+\gamma^2 R_{t+3}+\cdots+\gamma^{T-t-1} R_T \\
		&= R_{t+1}+\gamma G_{t+1}
	\end{aligned}
\end{equation}
In a round of Sanma, the reward for each action is not immediately received; instead, the cumulative reward is received once at the end of the round, as the score of the agent, with the intermediate rewards being zero. Therefore, we need a way to estimate the reward for each action, and we introduce a discount rate $0\leq\gamma\leq1$ to trade off the immediate and delayed reward: if $\gamma=0$, the agent is only concerned with maximizing the immediate rewards; as $\gamma$ approaches to 1, the cumulative reward takes future rewards into account more strongly, making the agent more farsighted. 

\begin{algorithm}[t]
	\caption{REINFORCE: Monte-Carlo Policy Gradient}
	\label{algo:reinforce}
	\begin{algorithmic}
		\Require Policy parametrization $\pi(a|s,\mathbf{w})$
		\Require Learning rate $\eta>0$
		\Require Discount rate $0\leq\gamma\leq1$
		\State Initialize policy parameter $\mathbf{w}$ from the pre-trained model
		\While{stopping criterion not met}
		\State Generate$\;S_0,A_0,R_1,...,S_{T-1},A_{T-1},R_T\;$from$\;\pi(\cdot|\cdot,\mathbf{w})$
		\State $R_1,...,R_{T-1}\gets0$, $R_T\gets\text{Round score/penalty}$
		\State $G_T\gets0$
		\ForAll{time step of the episode $t=T-1$ downto 0} 
		\State $G_t\gets R_{t+1}+\gamma G_{t+1}$
		\State $\mathbf{w}\gets\mathbf{w}+\eta\gamma^t G_t\nabla_\mathbf{w}\log\pi(A_t|S_t,\mathbf{w})$
		\EndFor
		\EndWhile
	\end{algorithmic}
\end{algorithm}

A policy $\pi$ is a mapping from each state to a probability distribution over actions, and defines the behaviour of the agent. Here, we use the pre-trained discard CNN as a parametrized policy to be improved through RL:
\begin{equation}
	\pi(a|s,\mathbf{w})=\Pr(A_t=a|S_t=s,\mathbf{w}_t=\mathbf{w})
\end{equation}
where $\mathbf{w}$ denotes the weights of the CNN. The state-value function $v_\pi(s)$ and the action-value function $q_\pi(s,a)$ for the policy can then be defined as follows:
\begin{align}
	v_\pi(s)&=\mathbb{E}_\pi[G_t|S_t=s] \\
	q_\pi(s,a)&=\mathbb{E}_\pi[G_t|S_t=s,A_t=a]
\end{align}
We can then define the performance measure $J(\mathbf{w})$ of our agent, given the initial state $s_0$ of a round of Sanma, and calculate its gradient
\begin{gather}
	J(\mathbf{w})=v_{\pi}(s_0) \propto \mathbb{E}_\pi\left[\sum_a q_\pi(S_t,a)\pi(a|S_t,\mathbf{w})\right]\\
	\begin{aligned}
		\nabla_\mathbf{w}J(\mathbf{w}) &\propto \mathbb{E}_\pi\left[\sum_a\pi(a|S_t,\mathbf{w})q_\pi(S_t,a)\frac{\nabla_\mathbf{w}\pi(a|S_t,\mathbf{w})}{\pi(a|S_t,\mathbf{w})}\right] \\
		&= \mathbb{E}_\pi\left[q_\pi(S_t,A_t)\frac{\nabla_\mathbf{w}\pi(A_t|S_t,\mathbf{w})}{\pi(A_t|S_t,\mathbf{w})}\right] \\
		&= \mathbb{E}_\pi\,\left[G_t\nabla_\mathbf{w}\log\pi(A_t|S_t,\mathbf{w})\right]
	\end{aligned}
\end{gather}

We then use the Monte Carlo policy gradient method to update the weights $\mathbf{w}$ as
\begin{equation}
	\mathbf{w}_{t+1}=\mathbf{w}_t+\eta\gamma^t G_t\nabla_\mathbf{w}\log\pi(A_t|S_t,\mathbf{w}_t)
\end{equation}

Algorithm~\ref{algo:reinforce} describes how we implement REINFORCE, the Monte Carlo policy gradient algorithm. After hyperparameter tuning, we adopt $\eta=10^{-3}$ and $\gamma=0.99$ as the optimum hyperparamter setting. Further details of hyperparameter tuning are decribed in Section~\ref{sec:experiments}. Since, for each state $S_t$, an action $A_t$ is sampled at random from the distribution defined by $\pi(A_t|S_t,\mathbf{w})$, this does introduce the risk of illegal discards: while ${\arg\max}_{A_t}\pi(A_t|S_t,\mathbf{w})$, which is used for actual predictions, has indeed always been in a player's private hand, the random sampling from $\pi(A_t|S_t,\mathbf{w})$ may still generate a choice that does not exist in the hand. Fortunately, this problem is very rare, occurring only twice during the entire training, and can be bypassed by simply skipping the problematic seeds.

\begin{figure}[t]
	\centerline{\includegraphics[width=0.667\columnwidth]{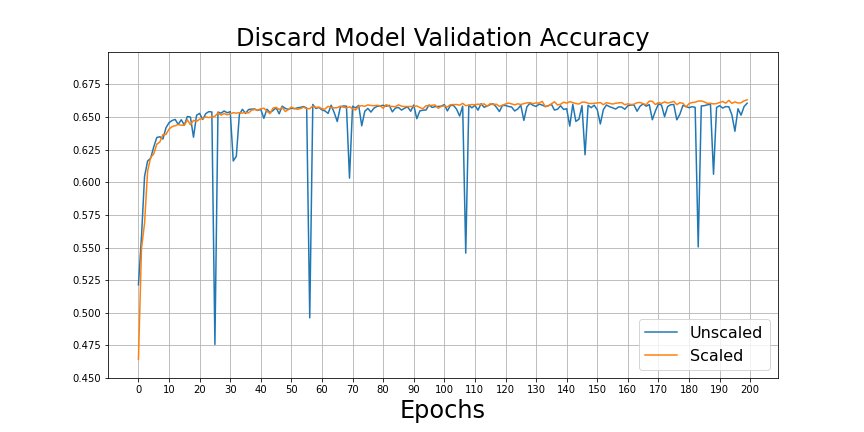}}
	\centerline{\includegraphics[width=0.5\columnwidth]{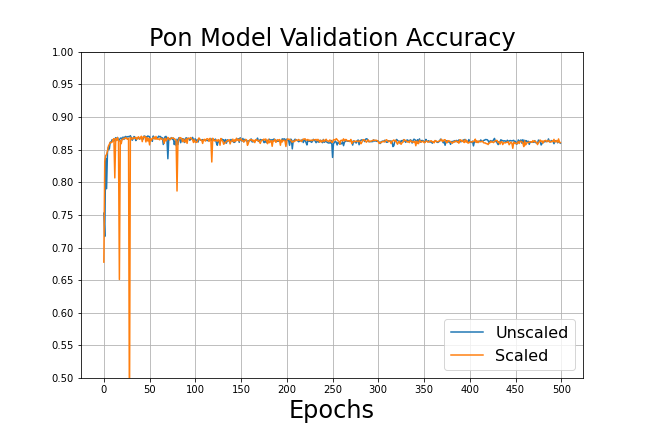}\includegraphics[width=0.5\columnwidth]{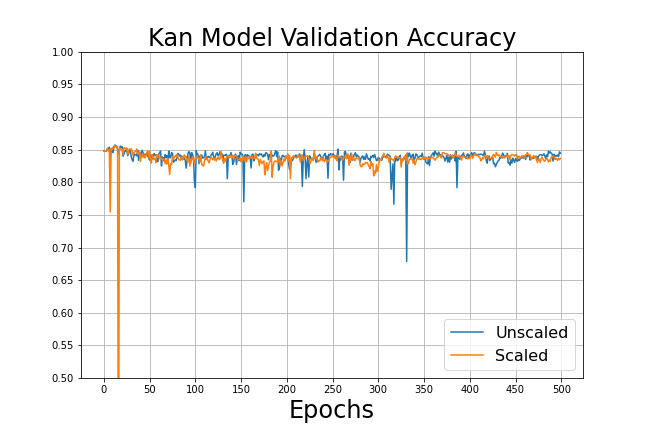}}
	\centerline{\includegraphics[width=0.5\columnwidth]{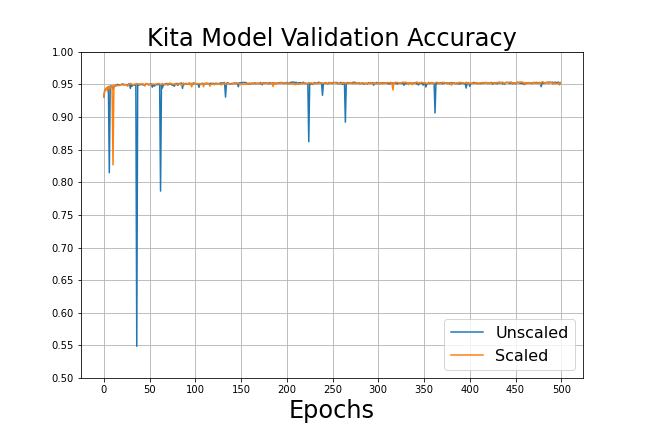}\includegraphics[width=0.5\columnwidth]{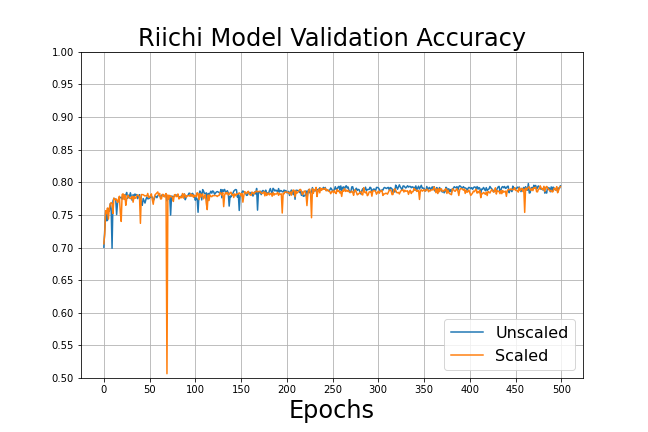}}
	\caption{Learning curves for the action models in supervised learning}
	\label{eval:learning-curves}
\end{figure}

\section{Experiments} \label{sec:experiments}

\subsection{Supervised Learning}

We sampled 50,000 rounds of Sanma game records from 2019 from the ``Houou'' (``phoenix'' in Japanese) table on Tenhou \cite{Tenhou} for training our models. 10\% of those data were divided, with stratification, to form the validation datasets of the actions. The ``Houou'' table is only open for the top 0.1\% of the ranked players, so its game records can be considered to be of high quality. To test the generalizability of our models, we sampled another 5,000 rounds from 2020 to form the test dataset. We use data from 2020, rather than 2019, as this potentially represents a harder generalization problem. The sizes of the datasets are shown in TABLE~\ref{table:dataset-sizes}. We also create a copy of each dataset, standardizing the data to zero mean and unit variance, to assess its effect on the models' performance. Our models are implemented using TensorFlow, and after hyperparameter tuning, Adam with learning rate $10^{-3}$ is used as the models' optimizer. The discard models, both with and without data standardization, are trained in mini-batches of size 64 for 200 epochs, and the rest of the models are trained in mini-batches of size 32 for 500 epochs. Each model is trained on 4 NVIDIA P100 GPUs with 64GB memory in total, and takes from 10 hours (Kan model) to 30 hours (discard model).

\begin{table}[t]
	\centering
	\caption{Sizes of the datasets}
	\resizebox{0.6\columnwidth}{!}{\begin{tabular}{|c|c|c|c|}
		\hline
		\multirow{2}{*}{\textbf{Action}} & \multicolumn{3}{c|}{\textbf{Dataset Size}} \\
		\cline{2-4}
		& \textbf{\textit{Training}} & \textbf{\textit{Validation}} & \textbf{\textit{Test}} \\
		\hline
		Discard & 797,285 & 88,588 & 147,444 \\
		Pon & 151,348 & 16,817 & 16,887 \\
		Kan & 34,319 & 3,814 & 3,548 \\
		Kita & 136,924 & 15,214 & 15,498 \\
		Riichi & 109,804 & 12,201 & 11,944 \\
		\hline
	\end{tabular}}
	\label{table:dataset-sizes}
\end{table}

\begin{table}[t]
	\centering
	\caption{Test accuracies for the action models in supervised learning}
	\resizebox{\columnwidth}{!}{\begin{tabular}{|c|c|c|c|c|}
			\hline
			\multirow{2}{*}{\textbf{Model}} & \multicolumn{4}{c|}{\textbf{Test Accuracy}} \\
			\cline{2-5}
			& \textbf{\textit{Meowjong}} & \textbf{\textit{Meowjong (scaled)}} & \textbf{\textit{Gao et al. \cite{Gao2018Mahjong}}} & \textbf{\textit{Suphx \cite{Li2020Suphx}}} \\
			\hline
			Discard & 65.81\% & 65.54\% & 68.8\% & 76.7\% \\
			Pon & 70.95\% & 72.10\% & 88.2\% & 91.9\% \\
			Kan & 92.45\% & 88.92\% & --- & 94.0\% \\
			Kita & 94.26\% & 94.44\% & --- & --- \\
			Riichi & 62.63\% & 64.29\% & --- & 85.7\% \\
			\hline
	\end{tabular}}
	\label{table:test-accuracy}
\end{table}

The learning curves of our models, showing mean validation accuracy can be seen in Fig.~\ref{eval:learning-curves}. They show a uniformly fast and satisfactory convergence. Compared to the discard models, the other models converged much more quickly, which is likely to be due to smaller dataset sizes. According to the learning curves, the scaled (data standardized) and unscaled models show no notable difference. Though one model may converge more stably than the other in some of the actions, this is not a general characteristic across all actions for either model.

TABLE~\ref{table:test-accuracy} shows the test accuracies of Meowjong's models, along with those achieved by previous works. Note that these test accuracies are not directly comparable due to the different training/validation/test data sources and structures, but they can serve as a rough reference. There is still no notable difference in the test accuracies between the unscaled and scaled models. In general, the test accuracies of Meowjong are very satisfactory, achieving Gao et al.'s level \cite{Gao2018Mahjong} at the most important action---discard. Although there is still a gap between Meowjong and Suphx, it is worth pointing out that Suphx used very large, 102/104-layer residual CNN structures, with training datasets of 4M--15M examples, and cost much more in computational power and time to train \cite{Li2020Suphx}, whereas Meowjong adopted a much simpler CNN structure, used a much smaller training dataset, and was much cheaper and faster to train.

\subsection{Reinforcement Learning}

\begin{figure}[t]
	\centerline{\includegraphics[width=0.8\columnwidth]{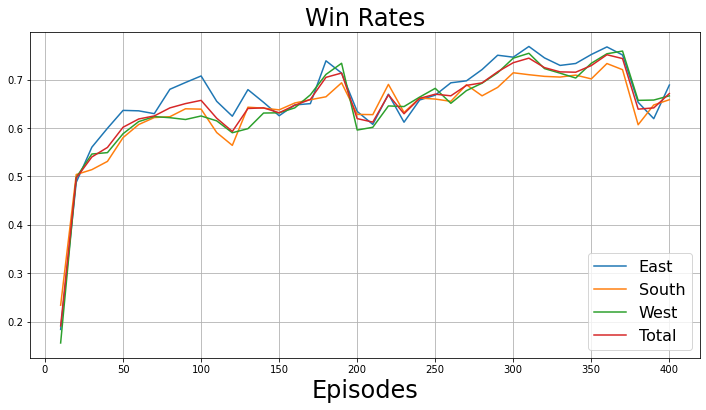}}
	\caption{Win rates of Meowjong in RL training}
	\label{fig:rl-win-rates}
\end{figure}

The discard model without data standardization was improved through self-play RL, using the REINFORCE (Monte Carlo policy gradient) algorithm, for 400 episodes. Evaluations against 2 baseline agents (described in Section \ref{sec:agents}) were carried out and recorded after every 10 episodes, each playing 500 rounds as East, South, and West. The curves of the win (1st place) rates are plotted in Fig.~\ref{fig:rl-win-rates}, showing a very successful improvement on all winds. The sudden drops on the win rates between Episodes 190/200 and Episodes 370/380 are probably due to the variance in policy updates, which are performed after every episode.

\begin{figure}[t]
	\centerline{\includegraphics[width=0.8\columnwidth]{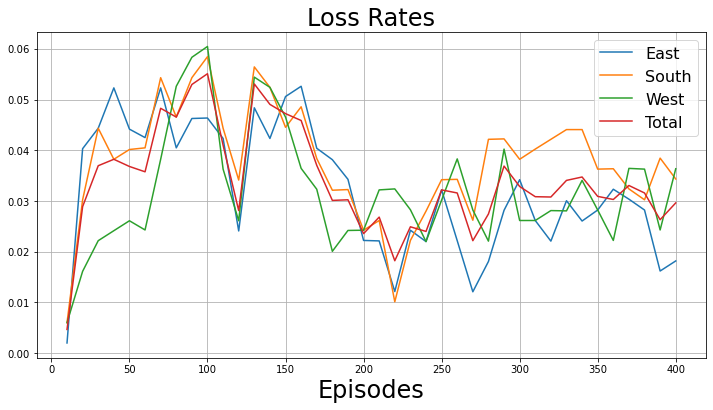}}
	\caption{Loss rates of the RL agent during training}
	\label{fig:rl-loss-rates}
\end{figure}

Fig.~\ref{fig:rl-loss-rates} shows the loss (3rd place) rates, which also gradually decrease over episodes, after sharp increases in the first 100 episodes. This suggests that the RL agent is likely to have learned an aggressive style, and finessed its skill through self-play, increasing its win rate while lowering its loss rate.

\begin{figure}[t]
	\centerline{\includegraphics[width=\columnwidth]{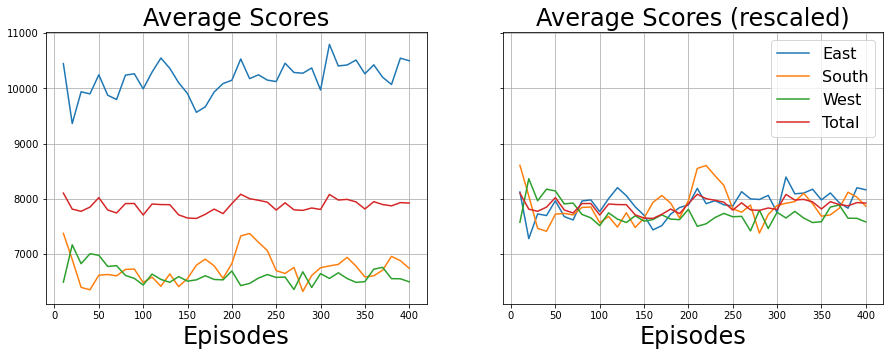}}
	\caption{Average scores of the RL agent during training}
	\label{fig:rl-avg-scores}
\end{figure}

Fig.~\ref{fig:rl-avg-scores} shows the average scores during RL training. The rescaled plots on the right remove the effect caused by the rule that the East player (the dealer) wins 50\% more, showing a consistent performance with no preference on any wind. The average scores are also stable across episodes, which is in accordance with the training target for Meowjong---maximising the winning probabilities rather than the round scores.

\begin{figure}[t]
	\centerline{\includegraphics[width=0.8\columnwidth]{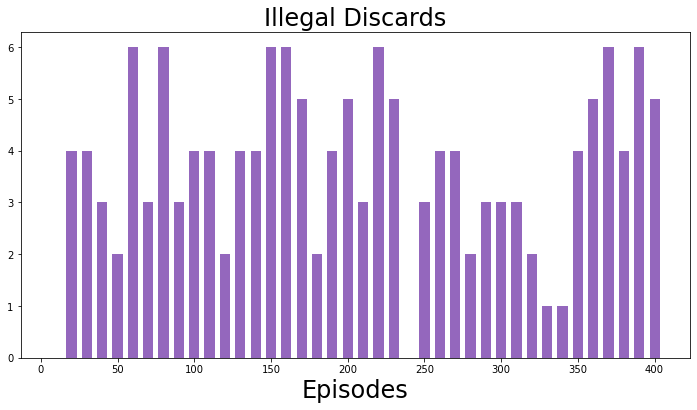}}
	\caption{Illegal discards of the RL agent}
	\label{fig:rl-illegal-discards}
\end{figure}

However, the RL training introduced a side-effect: the discard model may predict a tile to discard that does not exist in the hand. However, this is still very rare in evaluations, occurring at most 6 times in 1,500 rounds, and can be avoided in the deployment phase by predicting the legal tile with the highest probability. According to the bar plot in Figure \ref{fig:rl-illegal-discards}, we believe that the rate of illegal discards is independent of the episode.

\section{Agent Evaluation}

\subsection{Agents} \label{sec:agents}

We trained the following agents for evaluation: 
\begin{itemize}
	\item SL agent: the supervised learning agent with all 5 models trained using supervised learning;
	\item SL-scaled agent: the supervised learning agent using the data-standardized models;
	\item RL agent: the reinforcement learning agent with the Pon, Kan, Kita and Riichi models inherited from the SL agent, but its discard model initialized as the SL discard model and enhanced through RL.
\end{itemize}
We also built an agent that takes actions randomly to serve as a baseline. This agent is believed to have a similar power to the bots on the major online platforms.

\subsection{Evaluation Metrics}

The initial private tiles have large randomness and will greatly affect the win/loss of a game. In order to test whether one agent performs significantly better than another, we let the agents play a substantial number of rounds against each other, and conduct appropriate significance tests on the results. Since a round of Sanma can have 4 different outcomes: 1st place, 2nd place, 3rd place and draw, we assume that the game results amongst 3 agents can be modelled by a multinomial distribution $\text{Multinomial}(n,\mathbf{p})$, where:

\begin{itemize}
	\item $n$ is the total number of rounds played;
	\item $\mathbf{p}=\begin{bmatrix}
		p_1 & p_2 & p_3 & p_\text{draw}
	\end{bmatrix}$ denotes the probabilities of each outcome, with 1st place rate $p_1$, 2nd place rate $p_2$, 3rd place rate $p_3$, and draw rate $p_\text{draw}$.
\end{itemize}

The probability mass function for $\text{Multinomial}(n,\mathbf{p})$ is:
\begin{equation}
	\Pr(\mathbf{X}=\mathbf{x})=\frac{n!}{x_1!x_2!x_3!x_\text{draw}!}p_1^{x_1}p_2^{x_2}p_3^{x_3}p_\text{draw}^{x_\text{draw}}
\end{equation}
where $\mathbf{x}=\begin{bmatrix}
	x_1 & x_2 & x_3 & x_\text{draw}
\end{bmatrix}$ denotes the frequencies of the outcome, with 1st place frequency being $x_1$, 2nd place frequency being $x_2$, 3rd place frequency being $x_3$, and draw frequency being $x_\text{draw}$.

\begin{table}[t]
	\centering
	\caption{Comparisons between Meowjong agents vs. baseline agents}
	\resizebox{\columnwidth}{!}{\begin{tabular}{|c|c|r|r|r|r|}
			\hline
			\makecell[c]{\bf Agents\\ \bf (vs. Baseline)} & \textbf{Wind} & \makecell[r]{\bf 1st Place\\ \bf Rate} & \makecell[r]{\bf 2nd Place\\ \bf Rate} & \makecell[r]{\bf 3rd Place\\ \bf Rate} & \textbf{Draw Rate} \\
			\hline
			Baseline & --- & 0.02\% & 0.02\% & 0.02\% & 99.94\% \\
			\hline
			\multirow{4}{*}{SL} & East & 22.00\% & 0.06\% & 0.08\% & 77.86\% \\
			& South & 22.68\% & 0.06\% & 0.02\% & 77.24\% \\
			& West & 20.72\% & 0.16\% & 0.04\% & 79.08\% \\
			\cline{2-6}
			& Total & 21.80\% & 0.09\% & 0.05\% & 78.06\% \\
			\hline
			\multirow{4}{*}{SL-scaled} & East & 21.40\% & 0.04\% & 0.14\% & 78.42\% \\
			& South & 22.72\% & 0.08\% & 0.00\% & 77.20\% \\
			& West & 20.02\% & 0.20\% & 0.06\% & 79.72\% \\
			\cline{2-6}
			& Total & 21.38\% & 0.11\% & 0.06\% & 78.45\% \\
			\hline
			\multirow{4}{*}{RL} & East & 73.59\% & 0.02\% & 3.27\% & 23.12\% \\
			& South & 71.93\% & 0.08\% & 3.46\% & 24.53\% \\
			& West & 71.61\% & 0.06\% & 2.85\% & 25.48\% \\
			\cline{2-6}
			& Total & 72.38\% & 0.05\% & 3.19\% & 24.38\% \\
			\hline
	\end{tabular}}
	\label{table:agents-vs-baseline-results}
\end{table}

\begin{table}[t]
	\centering
	\caption{Significance test results for Meowjong vs. baseline agents}
	\resizebox{\columnwidth}{!}{\begin{tabular}{|c|c|r|r|r|r|r|r|}
			\hline
			\makecell[c]{\bf Agents\\ \bf (vs. Baseline)} & {\bf Wind} & $n$ & $x_1$ & $x_2$ & $x_3$ & $x_\text{draw}$ & $\Pr(\mathbf{X}\succeq\mathbf{x})$ \\
			\hline
			\multirow{4}{*}{SL} & East & 5,000 & 1,100 & 3 & 4 & 3,893 & $<O(\epsilon)^{\mathrm{a}}$ \\
			& South & 5,000 & 1,134 & 3 & 1 & 3,862 & $<O(\epsilon)$ \\
			& West & 5,000 & 1,036 & 8 & 2 & 3,954 & $<O(\epsilon)$ \\
			\cline{2-8}
			& Total & 15,000 & 3,270 & 14 & 7 & 11,709 & $<O(\epsilon)$ \\
			\hline
			\multirow{4}{*}{SL-scaled} & East & 5,000 & 1,070 & 2 & 7 & 3,921 & $<O(\epsilon)$ \\
			& South & 5,000 & 1,136 & 4 & 0 & 3,860 & $<O(\epsilon)$ \\
			& West & 5,000 & 1,001 & 10 & 3 & 3,986 & $<O(\epsilon)$ \\
			\cline{2-8}
			& Total & 15,000 & 3,207 & 16 & 10 & 11,767 & $<O(\epsilon)$ \\
			\hline
			\multirow{4}{*}{RL} & East & 4,949 & 3,642 & 1 & 162 & 1,144 & $<O(\epsilon)$ \\
			& South & 4,949 & 3,560 & 4 & 171 & 1,214 & $<O(\epsilon)$ \\
			& West & 4,949 & 3,544 & 3 & 141 & 1,261 & $<O(\epsilon)$ \\
			\cline{2-8}
			& Total & 14,847 & 10,746 & 8 & 474 & 3,619 & $<O(\epsilon)$ \\
			\hline
			\multicolumn{8}{l}{\makecell[l]{$^{\mathrm{a}}$ $\epsilon$ is the smallest representable positive \texttt{float64} value in Python, which equals\\$2^{-1074}\approx4.94\times10^{-324}$, and the constant factor is at most $15000^3$,  so the cu-\\mulative probability cannot exceed $1.67\times10^{-311}$.}}
	\end{tabular}}
	\vspace*{-\baselineskip}
	\label{table:agents-vs-baseline-test}
\end{table}

Before conducting the significance testing, it is necessary to define what is ``better''. Here we define ``better'' to be having both a higher 1st place rate and a lower 3rd place rate. If both the 1st place rates and the 3rd place rates are equal, the agent should prioritize on improving the draw rate, because 2nd place often comes from another player winning by self-draw or from the third player's discard, resulting in the agent themself having a zero or negative round score, and a guaranteed negative score difference. Two outcomes with one having higher values on both the 1st place rate and the 3rd place rate, or lower values on both the 1st place rate and the 3rd place rate, are not directly comparable. Therefore, for two outcomes $\mathbf{x}=\begin{bmatrix}
	x_1 & x_2 & x_3 & x_\text{draw}
\end{bmatrix}$ and $\mathbf{x}'=\begin{bmatrix}
	x'_1 & x'_2 & x'_3 & x'_\text{draw}
\end{bmatrix}$, we define the following strict partial order relation $\succ$, interpreted as ``an outcome is better than another'', to be
\begin{equation}
	\begin{aligned}
		\forall\mathbf{x},\mathbf{x}'.\ \mathbf{x}\succ\mathbf{x}'\iff &(x_1>x'_1\wedge x_3<x'_3)\ \vee \\
		&(x_1=x'_1\wedge x_3=x'_3\wedge x_\text{draw}>x'_\text{draw})
	\end{aligned}
\end{equation}

\begin{table}[t]
	\centering
	\caption{Comparisons between SL and RL agents}
	\resizebox{\columnwidth}{!}{\begin{tabular}{|c|c|r|r|r|r|}
			\hline
			\textbf{Agents} & \textbf{Wind} & \makecell[r]{\bf 1st Place\\ \bf Rate} & \makecell[r]{\bf 2nd Place\\ \bf Rate} & \makecell[r]{\bf 3rd Place\\ \bf Rate} & \textbf{Draw Rate} \\
			\hline
			\multirow{4}{*}{SL vs. SL} & East & 18.70\% & 11.90\% & 24.84\% & 44.56\% \\
			& South & 19.74\% & 24.32\% & 11.38\% & 44.56\% \\
			& West & 17.00\% & 19.22\% & 19.22\% & 44.56\% \\
			\cline{2-6}
			& Total & 18.48\% & 18.48\% & 18.48\% & 44.56\% \\
			\hline
			\multirow{4}{*}{SL vs. 2RL} & East & 5.76\% & 7.09\% & 81.53\% & 5.62\% \\
			& South & 6.10\% & 38.09\% & 49.36\% & 6.45\% \\
			& West & 4.90\% & 37.69\% & 51.68\% & 5.72\% \\
			\cline{2-6}
			& Total & 5.59\% & 27.62\% & 60.86\% & 5.93\% \\
			\hline
			\multirow{4}{*}{RL vs. 2SL} & East & 57.46\% & 7.75\% & 13.92\% & 20.87\% \\
			& South & 57.90\% & 16.34\% & 4.93\% & 20.83\% \\
			& West & 55.68\% & 18.18\% & 5.25\% & 20.89\% \\
			\cline{2-6}
			& Total & 57.02\% & 14.09\% & 8.03\% & 20.86\% \\
			\hline
	\end{tabular}}
	\vspace*{-\baselineskip}
	\label{table:rl-vs-sl-results}
\end{table}

\begin{table}[t]
	\centering
	\caption{Significance test results for RL vs. SL agents}
	\resizebox{\columnwidth}{!}{\begin{tabular}{|c|c|r|r|r|r|r|r|}
			\hline
			\textbf{Agents} & \textbf{Wind} & $n$ & $x_1$ & $x_2$ & $x_3$ & $x_\text{draw}$ & $\Pr(\mathbf{X}\succeq\mathbf{x})$ \\
			\hline
			\multirow{4}{*}{RL vs. 2SL} & East & 4,993 & 2,869 & 387 & 695 & 1,042 & $<O(\epsilon)$ \\
			& South & 4,993 & 2,891 & 816 & 246 & 1,040 & $<O(\epsilon)$ \\
			& West & 4,993 & 2,780 & 908 & 262 & 1,043 & $<O(\epsilon)$ \\
			\cline{2-8}
			& Total & 14,979 & 8,540 & 2,111 & 1,203 & 3,125 & $<O(\epsilon)$ \\
			\hline
			\multirow{4}{*}{SL vs. 2RL} & East & 4,996 & 288 & 354 & 4,073 & 281 & $>1-O(\epsilon)$ \\
			& South & 4,996 & 305 & 1,903 & 2,466 & 322 & $>1-O(\epsilon)$ \\
			& West & 4,996 & 245 & 1,883 & 2,582 & 286 & $>1-O(\epsilon)$ \\
			\cline{2-8}
			& Total & 14,988 & 838 & 4,140 & 9,121 & 889 & $>1-O(\epsilon)$ \\
			\hline
	\end{tabular}}
	\vspace*{-\baselineskip}
	\label{table:rl-vs-sl-test}
\end{table}

The same ``better'' relation also applies to the probabilities. Therefore, with the baseline probabilities $\mathbf{p}_0=\begin{bmatrix}
	p_1 & p_2 & p_3 & p_\text{draw}
\end{bmatrix}$ observed from the baseline game results, we set up the following hypotheses:
\begin{quote}
	Null hypothesis $H_0$: $\mathbf{p}=\mathbf{p}_0$ \\
	Alternative hypothesis $H_a$: $\mathbf{p}\succ\mathbf{p_0}$ \\
	Significance level: $\alpha=0.05$
\end{quote}
The cumulative distribution function of $\text{Multinomial}(n,\mathbf{p})$ can be used to test $H_0$, which can be calculated by the following formula:
\begin{equation}
	\begin{aligned}
	&\Pr(\mathbf{X}\succeq\mathbf{x}) = \sum_{\mathbf{x}'\succeq\mathbf{x}}\Pr(\mathbf{X}=\mathbf{x}') \\
	&= \sum_{x'_1>x_1}\sum_{x'_2}\sum_{x'_3<x_3}\Pr\left(\mathbf{X}=\left[x'_1\ x'_2\ x'_3\ (n-x'_1-x'_2-x'_3)\right]\right) \\
	&\ \ \ + \sum_{x'_2<x_2}\Pr\left(\mathbf{X}=\left[x_1\ x'_2\ x_3\ (n-x_1-x'_2-x_3)\right]\right) \\
	&\ \ \ + \Pr\left(\mathbf{X}=\left[x_1\ x_2\ x_3\ x_\text{draw}\right]\right)
	\end{aligned}
\end{equation}
where $\succeq$ is the non-strict version of $\succ$. If $\Pr(\mathbf{X}\succeq\mathbf{x})<\alpha$, then $H_0$ can be rejected. 

\subsection{Evaluation Results}

We simulated 5,000 rounds of Sanma games amongst 3 baseline agents, and 5,000 rounds for each of the SL, SL-scaled, and RL agents against 2 baseline agents, in each wind position. Their results, in terms of 1st/2nd/3rd-place and draw rates, are reported in TABLE~\ref{table:agents-vs-baseline-results}. The results clearly show that all Meowjong's trained agents can outperform a baseline agent. There is no significant difference in any of the rates between the SL and SL-scaled agents, and the RL agent's 1st place rates are also much larger than both SL agents. The RL agent also has higher 3rd-place rates, which suggests that it has learned an aggressive style through self-play.

\begin{figure}[t]
	\centerline{\includegraphics[width=\columnwidth]{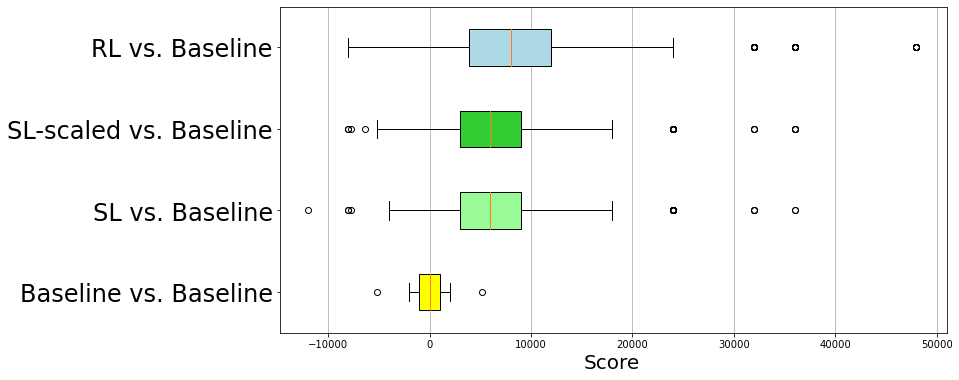}}
	\vspace*{-0.2\baselineskip}
	\caption{Scores of Meowjong agents against baseline agents}
	\vspace*{-\baselineskip}
	\label{fig:agents-vs-baseline}
\end{figure}

The results of the significance test are reported in TABLE~\ref{table:agents-vs-baseline-test}, which reject all $H_0$ and prove the significant difference. This is also confirmed by the box plots of all the scores shown in Fig.~\ref{fig:agents-vs-baseline}. Besides, it took on average 4.2 hours to simulate 5,000 rounds, which is about 3 seconds per round. This means Meowjong is able to make decisions very quickly, satisfying the time limit of all online platforms.

\subsection{Comparison Between SL and RL Agents}

\begin{figure}[t]
	\centerline{\includegraphics[width=\columnwidth]{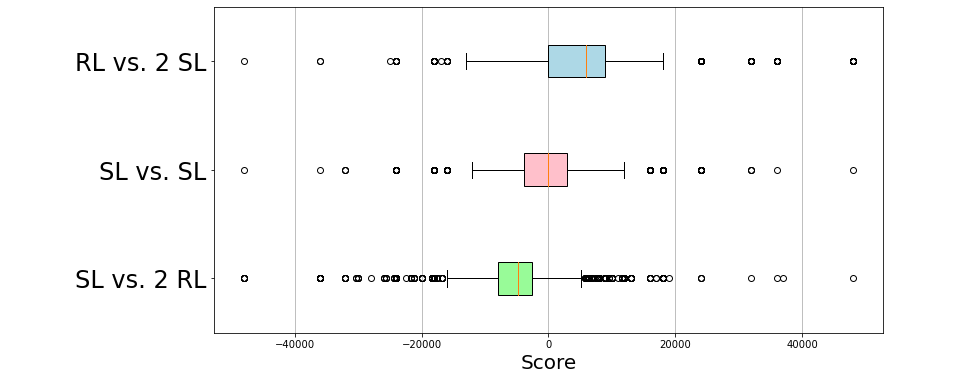}}
	\vspace*{-0.2\baselineskip}
	\caption{Scores of the SL agents vs. RL agents}
	\vspace*{-\baselineskip}
	\label{fig:rl-vs-sl}
\end{figure}

In this series of evaluations, we simulated 5,000 rounds amongst 3 SL agents, and 5,000 rounds in each wind positions in both SL vs. 2RL and RL vs. 2SL. The results are reported in TABLE~\ref{table:rl-vs-sl-results}. It is clear from the outcomes that an RL agent can outperform an SL agent. The higher 3rd place rates at the East position are due to the rule that the East player pays twice as much as the other player when a non-East player wins by self-draw. The significant difference can also be confirmed by the significance test results in TABLE~\ref{table:rl-vs-sl-test}, and the box plots of the scores in the simulation in Fig.~\ref{fig:rl-vs-sl}. Besides, according to the outcomes, both SL agents seem to perform better at the South position.

\section{Conclusions}

In this paper, we design a new data structure to encode the observable states in Sanma games, build an efficient CNN structure that solves the Sanma's decision-making problem, and train several Sanma agents using both supervised learning and reinforcement learning. All our action models achieve test accuracies comparable with AIs for 4-player Mahjong through supervised learning, and gain a significant further enhancement from reinforcement learning. Being the first ever AI in Sanma, we can claim that Meowjong stands as a state-of-the-art in this game.

In future work, we plan to take the multi-round ranking information into account, and let Meowjong try to maximize the full-game results, in addition to the single-round performance. An expert player adapts to different strategies flexibly based not only on their score and ranking at the beginning of each round, but also the progress of the entire game. For example, a player may play defensively in the later rounds when they possess a great lead, and may even lose deliberately to the lowest-placed player by a small amount in the last round, in order to secure the 1st place. If Meowjong could learn such a flexible full-game strategy, this would surely improve its multi-round performance even further. The nature of our data structure for encoding the game state already supports such an upgrade, and sequential neural network structures, such as GRUs and long short-term memory networks (LSTMs), might have the best potential for tackling this task.

\section*{Acknowledgment}

This work was performed using resources provided by the Cambridge Service for Data Driven Discovery (CSD3) operated by the University of Cambridge Research Computing Service, provided by Dell EMC and Intel using Tier-2 funding from the Engineering and Physical Sciences Research Council (capital grant EP/T022159/1), and DiRAC funding from the Science and Technology Facilities Council. For the purpose of open access, the authors have applied a Creative Commons Attribution (CC BY) licence to any Author Accepted Manuscript version arising. The code and data related to this publication are available at \url{https://github.com/VictorZXY/meowjong}.

\bibliographystyle{IEEEtran}
\bibliography{IEEEabrv, refs}

\end{document}